\documentclass{article}

\usepackage{microtype}
\usepackage{graphicx}
\usepackage{subcaption}
\usepackage{booktabs} 

\usepackage{hyperref}

\usepackage[preprint]{icml2026}

\usepackage{amsmath}
\usepackage{amssymb}
\usepackage{mathtools}
\usepackage{amsthm}

\usepackage{multirow}

\usepackage{algorithm}
\usepackage{algorithmic}

\usepackage[capitalize,noabbrev]{cleveref}

\usepackage{amsmath,amsfonts,bm}

\def\eqref#1{equation~\ref{#1}}

\def\1{\bm{1}}

\DeclareMathAlphabet{\mathsfit}{\encodingdefault}{\sfdefault}{m}{sl}
\SetMathAlphabet{\mathsfit}{bold}{\encodingdefault}{\sfdefault}{bx}{n}

\definecolor{bblue}{HTML}{1f77b4}
\definecolor{rred}{HTML}{d62728}
\definecolor{ggreen}{HTML}{2ca02c}
\definecolor{oorange}{HTML}{ff7f0e}
\definecolor{green}{rgb}{0.16, 0.67, 0.53}
\definecolor{blue}{rgb}{0.19, 0.55, 0.91}
\definecolor{red}{rgb}{0.8, 0.25, 0.33}
\definecolor{orange}{rgb}{0.91, 0.45, 0.32}
\definecolor{plum}{rgb}{0.867, 0.627, 0.867}
\definecolor{cyan}{rgb}{0.0, 1.0, 1.0}

\makeatletter
\newcommand\footnoteref[1]{\protected@xdef\@thefnmark{\ref{#1}}\@footnotemark}
\makeatother
\usepackage{graphicx}
\usepackage{xcolor}

\def\min{\mathop{\rm min}}
\def\max{\mathop{\rm max}}

\def\sup{\mathop{\rm sup}}
\def\inf{\mathop{\rm inf}}

\newcommand{\V}[1]{{{\boldsymbol #1}}}

\renewcommand{\1}{\V{1}}

\newcommand{\quoteIt}[1]{``#1''}

\newcommand{\essinf}{\mbox{essinf}}
\newcommand{\esssup}{\mbox{esssup}}

\renewcommand{\eqref}[1]{(\ref{#1})}

\newcommand{\EDmodified}[1]{{\color{red} #1}}
\newcommand{\EDcomments}[1]{{\EDmodified{Erick commented: #1}}}

\newcommand{\ACmodified}[1]{{\color{blue} #1}}
\newcommand{\ACcomments}[1]{{\ACmodified{AC commented: #1}}}

\newcommand{\removed}[1]{{}}
\newcommand{\moveToAppendix}[1]{{}}

\usepackage[textsize=tiny]{todonotes}
\usepackage{hyperref}

\usepackage{amsmath}
\usepackage{amssymb}
\usepackage{mathtools}
\usepackage{amsthm}

\theoremstyle{plain}
\newtheorem{theorem}{Theorem}[section]
\newtheorem{proposition}[theorem]{Proposition}
\newtheorem{lemma}[theorem]{Lemma}

\theoremstyle{definition}

\newtheorem{assumption}[theorem]{Assumption}
\theoremstyle{remark}

\usepackage[textsize=tiny]{todonotes}

\icmltitlerunning{Epistemic Robust Offline Reinforcement Learning}

\begin{document}

\twocolumn[
  \icmltitle{Epistemic Robust Offline Reinforcement Learning}



  \icmlsetsymbol{equal}{*}

  \begin{icmlauthorlist}
\icmlauthor{Abhilash Reddy Chenreddy}{hec}
\icmlauthor{Erick Delage}{hec}
\end{icmlauthorlist}

\icmlaffiliation{hec}{
GERAD, Department of Decision Sciences, HEC Montréal,  
Montréal, Québec H3T 2A7, Canada
}

\icmlcorrespondingauthor{Abhilash Reddy Chenreddy}{abhilash.chenreddy@hec.ca}

  \icmlkeywords{Machine Learning, ICML}

  \vskip 0.3in
]



\printAffiliationsAndNotice{}  

\begin{abstract}

Offline reinforcement learning learns policies from fixed datasets without further environment interaction. A key challenge in this setting is epistemic uncertainty, arising from limited or biased data coverage, particularly when the behavior policy systematically avoids certain actions. This can lead to inaccurate value estimates and unreliable generalization. Ensemble-based methods like SAC-N mitigate this by conservatively estimating Q-values using the ensemble minimum, but they require large ensembles and often conflate epistemic with aleatoric uncertainty.
To address these limitations, we propose a unified and generalizable framework that replaces discrete ensembles with compact uncertainty sets over Q-values. 
We also introduce a benchmark for evaluating offline RL algorithms under risk-sensitive behavior policies, and demonstrate that our method achieves improved robustness and generalization over ensemble-based baselines across both tabular and continuous state domains.

\end{abstract}

\section{Introduction}
\label{sec:introduction}

\removed{

Offline Reinforcement Learning (RL) aims to learn effective policies from static datasets without further interactions with the environment. A key challenge in this setting is that the uncertainty arises due to insufficient knowledge of the environment, particularly in regions of the state-action space that are poorly represented in the data. This is a prevalent problem in many real world applications where data collection is an inherently costly process. For instance, in personalized healthcare treatment planning or industrial control, collecting large scale interaction data may be impractical or unethical due to cost, safety, or privacy constraints (\cite{ghosh2022offline, levine2020offline}). This lack of coverage can lead to erroneous value estimates and unreliable generalization, particularly when standard RL algorithms attempt to extrapolate beyond observed data (\cite{yang2021believe}).

To mitigate this, modern offline RL algorithms such as Soft Actor-Critic with Ensembles (SAC-N) and its variants employ ensembles of Q-networks to quantify uncertainty of the Q-value estimates (\cite{an2021uncertainty}). These methods maintain a collection of $N$ independently initialized but jointly trained critics $\{Q^{(i)}_\theta\}_{i=1}^N$ and construct a conservative Bellman target using the pointwise minimum:

\begin{equation}
\label{eq:ensemble_target}
    y(s,a) := r + \gamma \min_{i \in [N]} Q^{(i)}_{\theta}(s', a') - \alpha \log \pi_\phi(a'|s'),
\end{equation}
where $(s,a,r,s') \sim \mathcal{D}$, with $\mathcal{D}$ as the dataset, and $a' \sim \pi_\phi(\cdot | s')$ is an action sampled from the policy $\pi_\phi$, parameterized by $\phi$, which maps a state $s'$ to a distribution over actions. Here, $\gamma \in (0,1]$ is the discount factor and $\alpha >0$ controls the entropy of the policy.
This formulation treats the minimum over ensemble members as a proxy for a lower confidence bound, promoting conservative estimates in uncertain regions. While empirically effective, ensemble based methods suffer from key limitations. First, reliable uncertainty estimation typically requires large ensemble sizes ($N \gg 1$), incurring substantial computational and memory overhead during both training and inference, and limiting scalability in high-dimensional domains(~\cite{wen2020batchensemble}). Second, the pointwise minimum discards inter-action correlations thereby reducing expressivity of the ensembles. Third,  ensembles conflate \emph{epistemic} and \emph{aleatoric} uncertainty, making it difficult to disentangle uncertainty due to data scarcity from inherent environmental stochasticity (~\cite{amini2020deep, osband2023epistemic}). This can hinder robust reasoning about what the agent does not know, and can lead to unsafe or overly conservative policies.

Even in scenarios with abundant data, epistemic uncertainty can persist due to behavioral policy bias, when the data is generated by a policy that systematically favors certain actions (\citep{schweighofer2022dataset}). To illustrate, consider a machine replacement problem (\cite{wiesemann2013robust}) formulated as a Markov decision process with \( \mathcal{S} = 10 \) states and \( \mathcal{A} = 2 \) actions. At each state \( s \in \{1, \ldots, 10\} \), the agent chooses either to continue operation (\( a = 1 \)) or to replace the machine (\( a = 2 \)). 
Continuing operation increases the chances of reaching a level of severe machine failure.  A \textit{risk-averse} behavioral policy may choose to replace early to minimize the chances of reaching the failure state, whereas a \textit{risk-seeking} policy may defer replacement until imminent failure becomes more certain to keep replacement costs to a minimum. 
Data collection under such fixed policies can result in certain severely underexplored state-action pairs. This sparse coverage leads to erroneous estimation of both the transition dynamics \( p(s' \mid s, a) \) and value function \( Q(s, a) \). The resulting epistemic uncertainty poses a significant challenge in offline reinforcement learning, where the agent must learn an optimal policy from static data without further environment interaction. Appendix~\ref{appendix:replacement-example} presents this example in detail, including the optimal policies under varying risk preferences and the resulting state-action visitation distributions under different risk tolerance levels. 

To overcome these limitations, we introduce a unified and generalizable alternative that replaces the discrete ensemble $\{Q^{(i)}(s, a)\}_{i=1}^N$ with a compact uncertainty set $\mathcal{U}(s) \subset \mathbb{R}^{|\mathcal{A}|}$ defined at each state. This leads to the following set-based Bellman target:
\begin{align}
\label{eq:set_backup}
y(s,a) := r + \gamma \min_{\mathbf{q} \in \mathcal{U}(s')} &\mathbb{E}_{a' \sim \pi_\phi(\cdot|s')} \left[ q(a') \right. \notag \\
&\left. - \alpha \log \pi_\phi(a'|s') \right]
\end{align}

where $\mathcal{U}(s')$ compactly models a set of plausible Q-value vectors over actions at state $s'$. This formulation enables richer representation of uncertainty. 

\vspace{0.5em}
\noindent
Our key contributions are as follows:
\begin{itemize}
    \item We propose the Epistemic Robust Soft Actor-Critic (ERSAC) model as an alternative and generalization for the ensemble based SAC-N method.  ERSAC exploits uncertainty sets to capture epistemic joint uncertainty about the Q-values of each action, thus enabling richer and more structured epistemic uncertainty modeling
    
    \item We integrate epistemic neural network (Epinets) (~\cite{osband2023epistemic}) in the new ERSAC framework and show how Epinets can be adapted to directly produce uncertainty set, circumventing the need for resampling at inference time. This latter implementation is shown to scale efficiently to high-dimensional offline RL settings. 
    
    \item We introduce a benchmark framework for evaluating offline RL algorithms under risk-sensitive behavioral policies, spanning both tabular and continuous state domains. Empirically, our method outperforms ensemble-based baselines across diverse tasks, achieving improved robustness and generalization.
\end{itemize}}

Offline Reinforcement Learning (RL) seeks to learn policies from static datasets without further environment interaction. A key challenge is epistemic uncertainty arising from poor state-action coverage leading to unreliable value estimates and unsafe extrapolation, especially in domains where data collection is expensive or risky (e.g., healthcare, industrial control)~\cite{ghosh2022offline, levine2020offline}. Standard RL algorithms may overgeneralize in these regions, leading to unreliable value estimates and poor policy performance~\cite{yang2021believe}. Ensemble-based methods like SAC-N address this by training multiple Q-networks and using a conservative Bellman target based on the pointwise minimum:
\begin{equation}
\label{eq:ensemble_target}
y(s,a) := r + \gamma \min_{i \in [N]} Q^{(i)}_{\theta}(s', a') - \alpha \log \pi_\phi(a'|s')
\end{equation}
where $(s,a,r,s') \sim \mathcal{D}$ is a sample from the offline dataset, and $a' \sim \pi_\phi(\cdot | s')$ is drawn from the stochastic policy $\pi_\phi$, parameterized by $\phi$, $\gamma \in (0,1]$ is the discount factor and $\alpha > 0$ governs the entropy regularization.

The ensemble based formulation treats the minimum as a proxy for a lower confidence bound, encouraging conservative value estimates in uncertain regions. While effective, this method has limitations. Large ensemble sizes ($N \gg 1$) are often needed for reliable uncertainty estimates, increasing computational and memory costs~\cite{wen2020batchensemble}. The minimum also ignores inter-action correlations, moreover, ensembles often conflate epistemic and aleatoric uncertainty~\cite{amini2020deep, osband2023epistemic}, making it difficult to distinguish model uncertainty from environment stochasticity, hindering robust and safe decision-making.

Epistemic uncertainty can persist even with large datasets when the behavior policy is biased. In the machine replacement problem~\cite{wiesemann2013robust}, where an agent decides whether to continue operating or replace a degrading machine across 10 states, a risk-averse policy may replace early to avoid failure, while a risk-seeking one may delay to reduce cost. These choices induce systematically different state-action coverage, leading to high epistemic uncertainty in underexplored regions~\cite{schweighofer2022dataset}. This issue is especially pronounced in offline RL, where no further interaction is possible to resolve uncertainty. Example discussed in Appendix~\ref{appendix:replacement-example} illustrates this with optimal and behavioral policies under different risk tolerances and the resulting coverage distributions.

To overcome these issues, we propose replacing the discrete ensemble $\{Q^{(i)}(s, a)\}_{i=1}^N$ with a compact uncertainty set $\mathcal{U}(s) \subset \mathbb{R}^{|\mathcal{A}|}$ defined per state. This yields a set-based Bellman target:
\begin{align}
y(s,a) :=\;& r 
\;+\; \gamma \min_{\mathbf{q} \in \mathcal{U}(s')}
\\[2pt] \nonumber
&\mathbb{E}_{a' \sim \pi_\phi(\cdot|s')}
   \big[ q(a') - \alpha \log \pi_\phi(a'|s') \big].
\label{eq:set_backup}
\end{align}
where $\mathcal{U}(s')$ represents plausible Q-value vectors over actions at state $s'$. This formulation enables a richer modeling of epistemic uncertainty, with improved sample efficiency and robustness. Our contributions can be described as:
\begin{itemize}
    \item We introduce ERSAC, a generalization of SAC-N using uncertainty sets to model structured epistemic uncertainty over Q-values.
    \item We integrate epistemic neural networks (Epinets)~\cite{osband2023epistemic} into ERSAC to directly produce uncertainty sets, removing the need for resampling.
    \item We develop a benchmark to evaluate offline RL under risk-sensitive behavior, demonstrating ERSAC's improved robustness and generalization across tasks.
\end{itemize}

For brevity, a detailed survey of related literature is deferred to Appendix~\ref{appendix:related-work}.

\section{Preliminaries}

We consider a Markov Decision Process (MDP) characterized by a possibly continuous state space $\mathcal{S}$, a discrete action space $\mathcal{A}$, a state-transition distribution $p(s_{t+1}|s_t, a_t)$, a reward function $r(s_t, a_t)$, and a discount factor $\gamma \in (0, 1)$. The reinforcement learning objective is to identify an optimal policy $\pi^*(\cdot|s)$, with $\pi^*(a|s)$ defining the likelihood of doing action $a$ when in state $s$, that maximizes the expected discounted cumulative reward $\mathbb{E}_{\pi}\left[ \sum_{t=0}^{\infty} \gamma^{t} r(s_t, a_t) \right]$.
Below, we summarize the Soft Actor-Critic (SAC) Algorithm and one of its adaptations for offline RL that performs conservative updates using an ensemble of Q-functions. 

\subsection{Soft actor critic (SAC)}The SAC framework optimizes the objective,
\[ J(\pi) = \mathbb{E}_{\pi} \left[ \sum_{t=0}^\infty \gamma^t \left( r(s_t, a_t) + \alpha \mathcal{H}(\pi(\cdot|s_t)) \right) \right],
\]
where $\mathcal{H}(\pi(\cdot|s)) = -\sum_{a \in \mathcal{A}} \pi(a|s) \log \pi(a|s)$ is the entropy of the policy, and $\alpha$ controls the trade-off between exploration and exploitation.

SAC employs parametric approximations for both the Q-function $Q_\theta(s, a)$ and the policy $\pi_\phi(a|s)$, which are updated using off-policy data from a replay buffer. The Q-function minimizes temporal-difference error, while the policy is optimized to maximize expected entropy-regularized Q-values, $\mathbb{E}_{a \sim \pi_\phi(\cdot|s)} \left[ Q_\theta(s,a) - \alpha \log \pi_\phi(a|s) \right]$. In this work, we use a discrete-action variant of SAC introduced in \cite{christodoulou2019soft}, and refer the reader to their work for implementation and theoretical details.

\removed{
\EDcomments{I think we can remove SAC}
\subsection{Soft Actor-Critic Algorithm (SAC)}
We adopt Soft Actor-Critic (SAC), originally developed for continuous action spaces, and adapt it to discrete actions (\cite{christodoulou2019soft}). By introducing entropy regularization, SAC strikes a balance between exploration and exploitation. Formally, for discrete actions, the SAC objective is:
\begin{equation}
    J(\pi) := \mathbb{E}_{\pi}\left[ \sum_{t=0}^{\infty} \gamma^{t} \left( r(s_t, a_t) + \alpha \mathcal{H}(\pi(\cdot|s_t)) \right) \right],
\end{equation}
where the entropy $\mathcal{H}(\pi(\cdot|s_t))$ is defined as

\begin{equation*}
    \mathcal{H}(\pi(\cdot|s_t)) := -\sum_{a \in \mathcal{A}} \pi(a|s_t) \log \pi(a|s_t),
\end{equation*}
and $\alpha$ is a temperature hyperparameter that controls the influence of the entropy term as a regularizer promoting policy stochasticity. We further define a Q-function $Q(s,a)$, estimating the entropy regularized expected cumulative reward from a state-action pair $(s,a)$ under policy $\pi$:

\begin{align}
Q(s,a) 
:= \mathbb{E}_{\pi} \Biggl[
\sum_{t=0}^{\infty} \gamma^{t} &\Bigl(r(s_t, a_t) + \alpha \mathcal{H}\bigl(\pi(\cdot \mid s_t)\bigr)\Bigr)  
\Biggr. \notag \\
&\Biggl. \Bigm| \; s_0 = s, \; a_0 = a
\Biggr]
\end{align}

We typically represent the Q-function using a parametric model $Q_\theta(s,a)$, e.g. a neural network, to effectively handle continuous and high-dimensional state spaces. The policy $\pi_\phi(a|s)$, parameterized by $\phi$, is defined as a categorical probability distribution over discrete actions conditioned on continuous states, facilitating straightforward computation of entropy terms.

The parameters of the policy and Q-functions are updated iteratively using off-policy experiences drawn from a replay buffer $\mathcal{D}$. Specifically, the Q-function parameters $\theta$ are updated to minimize the temporal difference (TD) error:
\begin{align*}
\theta \leftarrow \theta 
&- \eta_Q \nabla_\theta \,
\mathbb{E}_{(s, a, r, s') \sim \mathcal{D}} 
\Bigl[ \bigl( Q_\theta(s,a) - \\
&\bigl( r + \gamma \, 
\mathbb{E}_{a' \sim \pi(\cdot \mid s')} \bigl[ 
Q_{\theta'}(s',a') - \alpha \log \pi_\phi(a' \mid s') \bigr] 
\bigr) \bigr)^2 \Bigr]
\end{align*}

where $\eta_Q$ is the Q-function learning rate, and $\theta'$ denotes parameters of a target Q-network periodically synchronized with $\theta$ to enhance training stability. The policy parameters $\phi$ are updated to maximize the entropy-regularized expected Q-values:
\begin{equation*}
    \phi \leftarrow \phi + \eta_\pi \nabla_\phi \mathbb{E}_{s \sim \mathcal{D}, a \sim \pi_\phi(\cdot \mid s)} [Q_\theta(s,a) - \alpha \log \pi_\phi(a|s)],
\end{equation*}
where $\eta_\pi$ is the policy learning rate.
}

\subsection{SAC with an Ensemble of Q-functions (SAC-N)}
\label{SACN}
While SAC provides a stable framework for policy learning, applying it to offline RL is challenging since the agent relies solely on a fixed dataset. This makes SAC susceptible to overestimation bias, where the Q-function extrapolates inaccurately to out-of-distribution state-action pairs. Such bias is problematic during policy improvement, which favors actions with high Q-values, potentially leading to unsafe or suboptimal behavior. To mitigate this, \citet{an2021uncertainty} proposed SAC-N, which uses an ensemble of $N$ Q-functions \(\{Q_{\theta_i}\}_{i=1}^N\)  to capture epistemic uncertainty and reduce overestimation. Each $Q{\theta_i}$ estimates expected return, and a target ensemble \(\{Q_{\theta_i'}\}_{i=1}^N\) is updated via Polyak averaging. The Q-function update adopts a clipped double Q-learning–style target \citep{fujimoto2018addressing}, extended in SAC-N by taking the minimum over the ensemble:
\begin{align}
y(r, s',a') :=  \;r + \gamma \left( \min_i Q_{\theta_i'}(s', a') - \alpha \log \pi_\phi(a' \mid s') \right)
\label{eq:targetSAC}
\end{align}

Using the minimum over the ensemble provides a conservative estimate of the expected return, reducing propagation of overestimated values from out-of-distribution state-action pairs common in offline datasets. Each Q-function \( Q_{\theta_i} \) is updated by minimizing the mean squared Bellman error between its prediction and the target \( y(r, s', a') \):
\begin{equation}
\label{eq:ensemble_q_loss}
\begin{aligned}
\mathcal{L}_{Q}(\theta_i) :=\ 
\mathbb{E}_{\substack{(s,a,r,s') \sim \mathcal{D},\\ a' \sim \pi_\phi(\cdot \mid s')}} 
\biggl[ \Bigl( Q_{\theta_i}(s, a) 
- y(r, s', a') \Bigr)^2 \biggr]
\end{aligned}
\end{equation}
where \(\mathcal{D}\) denotes the static replay buffer of environment interactions, which, unlike in online RL, is fixed and is collected a priori without further interactions. The policy $\pi_\phi$ is then optimized to maximize the conservative estimate of the expected return (minimum Q-value across the ensemble) while incorporating the entropy regularization term:
\begin{equation}
\label{eq:ensemble_pi_loss}
\begin{aligned}
\mathcal{J}_{\pi}(\phi) :=\ 
\hspace{-0.1cm}\mathbb{E}_{\substack{s \sim \mathcal{D},\\ a \sim \pi_\phi(\cdot \mid s)}} 
\biggl[ \min_{i} Q_{\theta_i}(s, a) 
- \alpha \log \pi_\phi(a \mid s) \biggr]
\end{aligned}
\end{equation}
This objective balances maximizing a conservative estimate of expected returns with encouraging high entropy, which promotes stochastic action selection. Greater entropy helps the policy explore beyond frequent actions in the offline dataset, particularly useful early in training to avoid overfitting to spurious correlations. Following \cite{haarnoja2018soft}, the entropy coefficient \( \alpha \) is learned by minimizing a dual objective that aligns policy entropy with a target value, allowing the agent to maintain high entropy under uncertainty and gradually shift toward reward maximization.

Although SAC-N mitigates overestimation by maintaining an ensemble of Q-functions, it often requires a large ensemble size for stable performance. To address this, 
\cite{an2021uncertainty} introduced the \textbf{Ensemble-Diversified Actor-Critic (EDAC)},  which adds a diversification term to encourage diversity among the 
Q-function ensemble members. In continuous action setting, they quantify similarity using an ensemble similarity (ES) metric defined as:
\begin{equation*}
\frac{\langle\nabla_a Q_{\theta_i}(s, a), \nabla_a Q_{\theta_j}(s, a)\rangle}{\|\nabla_a Q_{\theta_i}(s, a)\| \|\nabla_a Q_{\theta_j}(s, a)\|},
\end{equation*}
which measures the cosine similarity between the gradients of different Q-functions with respect to the action vector. In the discrete action setting, where \(\nabla_a Q(s, a)\) is ill defined, we adapt the ES metric by instead computing the mean squared deviation between the Q-values across all actions. Specifically, we define $g_{\theta}(s,a) := \big(Q_{\theta}(s,a') - Q_{\theta}(s,a)\big)_{a' \in \mathcal{A}},$ and compute the cosine similarity between $g_{\theta_i}(s,a)$ and $g_{\theta_j}(s,a)$:

\begin{equation}
\begin{aligned}
\text{ES}_{\theta_i,\theta_j}(s,a)
&:= 
\frac{
\sum_{a' \in \mathcal{A}}
\Delta_i(a')\,\Delta_j(a')
}{
\sqrt{
\sum_{a' \in \mathcal{A}} \Delta_i(a')^2
}
\,
\sqrt{
\sum_{a' \in \mathcal{A}} \Delta_j(a')^2
}
},\\
\end{aligned}
\end{equation}
where $\Delta_k(a') := Q_{\theta_k}(s,a') - Q_{\theta_k}(s,a)$. The diversification loss is then given by:
\begin{equation*}
\mathcal{L}_{\text{ES}}(\theta) := 
\mathbb{E}_{(s,a) \sim \mathcal{D}}\!\left[\sum_{i=1}^N \sum_{j=i+1}^N \text{ES}_{\theta_i, \theta_j}(s,a)\right].
\end{equation*}
where $\theta$ is short for $\{\theta_i\}_{i=1}^N$. 
The overall loss for each Q-function incorporates this diversification term:
\begin{equation}
\bar{\mathcal{L}}_{Q}(\theta) := (1/N)\sum_{i=1}^N\mathcal{L}_{Q}(\theta_i) + \eta \mathcal{L}_{\text{ES}}(\theta),
\end{equation}
where $\eta$ is a hyperparameter controlling the strength of the diversity regularization. Encouraging diversity among the Q-functions was shown empirically to improve uncertainty estimation and leads to more reliable policy learning.

\section{Epistemic Robustness with SAC}  
\label{sec:ERSAC}
We start by formalizing the uncertainty captured by such an ensemble by modeling the long term actions values at a given state \(s\) as a distribution \(F_\theta^q(s) \in \mathcal{M}(\mathbb{R}^{|\mathcal{A}|})\). Here, \(F_\theta^q(s)\) defines a probability measure over Q-value vectors \(q \in \mathbb{R}^{|\mathcal{A}|}\), induced by the variability among the Q-functions, and  parameterized through \(\theta\).  Each sample \( \tilde{q} \sim F^q_\theta(s) \) is a vector in \( \mathbb{R}^{|\mathcal{A}|} \) representing the epistemic uncertainty about the action-wise values 
\( Q(s, \cdot) \).  
For example, in the case of SAC-N, this distribution takes the form of a scenario-based distribution:
\begin{equation}
F^q_\theta(s) := \frac{1}{N} \sum_{i=1}^N \delta_{Q_{\theta_i}(s,\cdot)},\label{eq:ensembleF}
\end{equation}
where \(\delta_x\) is the Dirac measure centered at \(x \in \mathbb{R}^{|\mathcal{A}|}\). Given a Q-value distribution \( F^q_\theta: \mathcal{S} \to \mathcal{M}(\mathbb{R}^{|\mathcal{A}|}) \), mapping each state \(s \in \mathcal{S}\) to a probability measure over Q-value vectors, we define an uncertainty set operator,
\[
\mathcal{U}: \mathcal{M}(\mathbb{R}^{|\mathcal{A}|}) \to \mathcal{C}(\mathbb{R}^{|\mathcal{A}|}),
\]
that maps a Q-value distribution to a compact set of plausible Q-value vectors. The composition \(\mathcal{U} \circ F^q_\theta: \mathcal{S} \to \mathcal{C}(\mathbb{R}^{|\mathcal{A}|})\) defines an epistemic uncertainty set \(\mathcal{U}(F^q_\theta(s))\) in each state \(s\), which can be used to construct robust evaluation and optimization of policies. For notational simplicity, we will use \( \mathcal{U}_\theta(s) \) as shorthand for \( \mathcal{U}(F^q_{\theta}(s)) \) when the dependencies on \( F^q_{\theta} \) are clear from context.

In the next section, we introduce our proposed framework, \textbf{Epistemic Robust Soft Actor-Critic (ERSAC)}, which generalizes SAC-N by leveraging uncertainty sets derived from Q-value distributions. We first present an ensemble-based version of ERSAC and highlight its connection to SAC-N. We then formalize the algorithm, detailing its key components,  the set-based Bellman backup and the robust policy update.

\subsection{The Epistemic Robust SAC (ERSAC) Model}
\removed{
\EDcomments{I suggest something like \quoteIt{Epistemic Robust SAC (ERSAC)}}
\ACcomments{I am thinking the following naming system\\
Epistemic-Robust SAC with Box set from N ensembles: ERSAC-B(N=x)\\
Epistemic-Robust SAC with Convex Hull set from N ensembles: ERSAC-C(N=x)\\
Epistemic-Robust SAC with Ellipsoid from N ensembles: ERSAC-E(N=x)\\
Epistemic-Robust SAC with Ellipsoid from epinet : ERSAC-E(Epi)}
\EDcomments{I think ERSAC-X-Y would be fine, where X in {B,CH,E} and Y in {N,Epi}. This way when Y is absent it just means a generic sampling operator. I dont know what you mean with $N=x$}
}

As in SAC-N, ERSAC trains the Q-function by minimizing the expected squared Bellman error between a sampled realization and a conservative target derived from the Q-distribution \( F^q_\theta \). Specifically, for each next state \( s' \in \mathcal{S} \), the target in \eqref{eq:targetSAC} is modified to:
\begin{equation}
\label{eq:rob_target}
\begin{aligned}
y(r, s') :=\;&
r + \gamma \Biggl(
\min_{q \in \mathcal{U}(F^q_{\theta'}(s'))}
\mathbb{E}_{a' \sim \pi_\phi(\cdot \mid s')}
\bigl[
q(s', a') \\
& \qquad \qquad - \alpha \log \pi_\phi(a' \mid s')
\bigr]
\Biggr)
\end{aligned}
\end{equation}
where the minimum operator
provides a robust estimate of the regularized expected total discounted return. 
We refer the reader to \cite{ben2015deriving} for closed form expressions of $\min_{q \in \mathcal{U}} \langle v, q \rangle$ for a list of popular forms of uncertainty sets. The loss function in \eqref{eq:ensemble_q_loss}  is then redefined as:
\begin{equation}
\label{eq:distributional_q_loss}
\mathcal{L}_Q^{R}(\theta) := \mathbb{E}_{(s,a,r,s') \sim \mathcal{D},\ \tilde{q} \sim F^q_\theta(s)} \left[ \left( \tilde{q}(a) - y(r,s') \right)^2 \right].\footnote{It is important to note that without additional regularization, the objective in \eqref{eq:distributional_q_loss} may admit a degenerate solution \(F_{\theta^*}^q(s) = \delta_{\bar{q}(s,\cdot)}\), where \(\bar{q}(s,a) := \mathbb{E}_{(s,a,r,s') \sim \mathcal{D}}[y(r,s')]\), which collapses the distribution to a deterministic point estimate. In practice, this requires regularization strategies 
such as early stopping, entropy constraints on \(F^q_\theta\), or prior-based regularization to avoid mode collapse.}
\end{equation}

Similar to the Q-value target, the policy loss in the epistemic robust setting replaces the ensemble minimum with a worst-case expectation over the uncertainty set. The robust policy loss \eqref{eq:ensemble_pi_loss} becomes $\mathcal{J}^{R}_{\pi}(\phi)$,
\begin{align}
:&=\label{eq:policy_loss_robust} \mathbb{E}_{s \sim \mathcal{D}} \biggl[ 
\min_{q \in \mathcal{U}_\theta(s)} 
\mathbb{E}_{a \sim \pi_\phi(\cdot \mid s)} 
\Bigl[ q(a) - \alpha \log \pi_\phi(a \mid s) \Bigr] 
\biggr] \\
&= \mathbb{E}_{\substack{s \sim \mathcal{D},\\ a \sim \pi_\phi(\cdot \mid s)}} 
\biggl[ \min_{q \in \mathcal{U}_\theta(s)} 
\langle \pi_\phi(\cdot \mid s), q \rangle 
- \alpha \log \pi_\phi(a \mid s) \biggr]\notag
\end{align}

Importantly, when using an ensemble based representation, the ERSAC formulation encompasses SAC-N as a special case under a particular choice of uncertainty set. We formalize this connection in the following proposition and defer the proof to Appendix~\ref{appendix:sacn-special-case}.

\begin{proposition}
\label{lem:box_set}
Let \( F_\theta^q(s) \) be defined as in Equation~\eqref{eq:ensembleF}, and let the uncertainty set operator be defined as
\begin{equation}
\label{eq:box_set_def}
\mathcal{U}_{\text{box}}(F_\theta^q(s)) := 
\mathop{\times}_{a \in \mathcal{A}} \left[
\mathop{\essinf}\limits_{\substack{\tilde{q} \sim F_\theta^q(s)}} [\tilde{q}(a)],\ 
\mathop{\esssup}\limits_{\substack{\tilde{q} \sim F_\theta^q(s)}} [\tilde{q}(a)]
\right]
\end{equation}

i.e., a coordinate-wise box containing the support of \( F_\theta^q(s) \), under which the robust losses reduce to those of SAC-N: \( \mathcal{L}^{R}_Q(\theta) = \tfrac{1}{N} \sum_{i=1}^N \mathcal{L}_Q(\theta_i) + C \) and \( \mathcal{J}_\pi^{R} = \mathcal{J}_\pi \), for some constant \( C \in \mathbb{R} \) independent of \( \theta \).
\end{proposition}

\removed{

\begin{proof}

We begin by analysing the robust estimator term present in both the conservative target value \eqref{eq:rob_target} and the policy loss \eqref{eq:policy_loss_robust}: $\min_{q \in \mathcal{U}_\theta(s)} \langle \pi_\phi(\cdot \mid s), q \rangle$.
Given that the uncertainty set is defined as a coordinate-wise product box and that $\pi_\phi(\cdot \mid s)\geq 0$, the minimum must be achieved at the coordinate-wise lower bound:
\[
q^*(a) = \essinf_{\tilde{q}\sim F_\theta^q(s)}[\tilde{q}(a)] = \essinf_{\tilde{i}\sim U(N)}[Q_{\theta_{\tilde{i}}}(s,a)] =\min_{i \in [N]} Q_{\theta_i}(s,a), \quad \forall a \in \mathcal{A}.
\]
The robust evaluation then becomes,
\begin{align*}
\min_{q \in \mathcal{U}_\theta(s)} \langle \pi_\phi(\cdot \mid s), q \rangle = \sum_{a \in \mathcal{A}} \pi_\phi(a \mid s) \min_{i \in [N]} Q_{\theta_i}(s,a) = \mathbb{E}_{a \sim \pi_\phi(\cdot \mid s)} \left[ \min_{i \in [N]} Q_{\theta_i}(s,a) \right]
\end{align*}

Hence, the conservative target value becomes
\begin{align*}
       y(r,s') &= r + \gamma \bigg( 
\mathbb{E}_{a'\sim\pi_\phi(\cdot \mid s')}\left[\min_{i \in [N]} Q_{\theta_i}(s',a') - \alpha  \log \pi_\phi(a' \mid s') \right] \\
       &=\mathbb{E}_{a' \sim \pi_\phi(\cdot \mid s')} 
\left[r + \gamma \bigg( 
\min_{i \in [N]} Q_i(s',a') - \alpha \,  \log \pi_\phi(a' \mid s') \right] \\
&= \mathbb{E}_{a' \sim \pi_\phi(\cdot \mid s')} 
\left[y(r,s',a')\right]
   \end{align*}
We further have that
\begin{align*}
    \mathcal{L}_Q^R(\theta)&=\mathbb{E}_{(s,a,r,s') \sim \mathcal{D},\ \tilde{q} \sim F^q_\theta(s)} \left[ \left( \tilde{q}(a) - y(r,s') \right)^2 \right]\\
    &=\mathbb{E}_{(s,a,r,s') \sim \mathcal{D},\ \tilde{q} \sim F^q_\theta(s)} \left[ \left( \tilde{q}(a) - \mathbb{E}_{a' \sim \pi_\phi(\cdot \mid s')}[y(r,s',a')]\right)^2 \right] \\
    &=\mathbb{E}_{(s,a,r,s') \sim \mathcal{D},\ \tilde{q} \sim F^q_\theta(s)} \left[ \tilde{q}(a)^2 - 2\tilde{q}(a)\mathbb{E}_{a' \sim \pi_\phi(\cdot \mid s')}[y(r,s',a')] + \mathbb{E}_{a' \sim \pi_\phi(\cdot \mid s')}[y(r,s',a')]^2 \right] \\
    &=\mathbb{E}_{(s,a,r,s') \sim \mathcal{D},\ \tilde{q} \sim F^q_\theta(s)} \left[ \tilde{q}(a)^2 - 2\tilde{q}(a)\mathbb{E}_{a' \sim \pi_\phi(\cdot \mid s')}[y(r,s',a')] + \mathbb{E}_{a' \sim \pi_\phi(\cdot \mid s')}[y(r,s',a')^2] \right] \\
    &\qquad + \mathbb{E}_{(s,a,r,s') \sim \mathcal{D}} \left[\mathbb{E}_{a' \sim \pi_\phi(\cdot \mid s')}[y(r,s',a')]^2 - \mathbb{E}_{a' \sim \pi_\phi(\cdot \mid s')}[y(r,s',a')^2]\right]\\
    &=\mathbb{E}_{(s,a,r,s') \sim \mathcal{D},\ \tilde{q} \sim F^q_\theta(s),a' \sim \pi_\phi(\cdot \mid s')} \left[ \tilde{q}(a)^2 - 2\tilde{q}(a)y(r,s',a') + y(r,s',a')^2 \right] + C\\    
    &= \mathbb{E}_{(s,a,r,s') \sim \mathcal{D},\ \tilde{q} \sim F^q_\theta(s),a' \sim \pi_\phi(\cdot \mid s')} \left[ \left( \tilde{q}(a) - y(r,s',a')\right)^2 \right] + C \\
    &=(1/N) \sum_i \mathbb{E}_{(s,a,r,s') \sim \mathcal{D},a' \sim \pi_\phi(\cdot \mid s')} \left[ \left( Q_{\theta_i}(s, a) - y(r,s',a')\right)^2 \right] + C \\
    &=(1/N) \sum_i \mathcal{L}_Q(\theta_i) + C 
\end{align*}
where
\begin{align*}
C&:=\mathbb{E}_{(s,a,r,s') \sim \mathcal{D}}[(\mathbb{E}_{a' \sim \pi_\phi(\cdot \mid s')}[y(r,s',a')])^2] - \mathbb{E}_{(s,a,r,s') \sim \mathcal{D},a' \sim \pi_\phi(\cdot \mid s')} \left[y(r,s',a')^2\right]    
\end{align*}
due to $\tilde{q}(a)$ being independent of $y(r,s',a')$ given $(s,a,r,s')$.

On the other hand, we have that:
\begin{align*}
\mathcal{J}_{\pi}^R(\phi)&=\mathbb{E}_{s \sim \mathcal{D},a \sim \pi_\phi(\cdot \mid s)} \bigg[ 
\min_{q \in \mathcal{U}_\theta(s)} \langle \pi_\phi(\cdot \mid s), q \rangle  - \alpha \log \pi_\phi(a \mid s) 
\bigg]\\
&=\mathbb{E}_{s \sim \mathcal{D},a \sim \pi_\phi(\cdot \mid s)} \bigg[ 
\mathbb{E}_{a'\sim\pi_\phi(\cdot \mid s)}[\min_{i \in [N]} Q_{\theta_i}(s,a')]  - \alpha \log \pi_\phi(a \mid s) 
\bigg]\\
&=\mathbb{E}_{s \sim \mathcal{D},a \sim \pi_\phi(\cdot \mid s)} \bigg[ 
\min_{i \in [N]} Q_{\theta_i}(s,a)  - \alpha \log \pi_\phi(a \mid s) 
\bigg]\\
&=\mathcal{J}_\pi(\phi).
\end{align*}

\end{proof}}

This result demonstrates that ERSAC generalizes SAC-N under a unified uncertainty set framework. In the next section, for an arbitrary compact set representation $\mathcal{U}_\theta(s)$, we outline the detailed training algorithm.

\subsection{The ERSAC Training Algorithm}
\label{sec:training_algo}

Previously, we modeled \(F^q_\theta(s)\) such that each sample \(\tilde{q} \sim F^q_\theta(s)\) is a Q-value vector in \(\mathbb{R}^{|\mathcal{A}|}\), representing \(Q(s, \cdot)\). To generalize this, we adopt the reparameterized formulation from Assumption~\ref{ass:reparamTrick}. 
\begin{assumption}\label{ass:reparamTrick}
    $F_\theta^q$ is associated to a sampling operator $\mathfrak{q}_\theta(s,a,z)$ and a distribution  $F_z \in \mathcal{M}(\mathbb{R}^{d_z})$,    such that $\mathfrak{q}_\theta(s,\cdot,\tilde{z})$ follows $F_\theta^q(s)$ when $\tilde{z}\sim F_z$.
\end{assumption}
Given a noise sample \( \tilde{z} \sim F_z \), a corresponding Q-vector sample \( \tilde{q} \sim F^q_\theta(s) \) is obtained by evaluating the sampling operator over all actions:
\[
\tilde{q}(a) := \mathfrak{q}_\theta(s, a, \tilde{z}), \quad \text{for all } a \in \mathcal{A}.
\]

This reparameterization generalizes the ensemble model in Equation~\ref{eq:ensembleF} as a special case, where the latent variable \( \tilde{z} \in \{1, \ldots, N\} \) indexes a finite set of Q-functions, and \( q_\theta(s, a, \hat{z}) = Q_{\theta_{\tilde{z}}}(s, a) \).

In order to minimize $\mathcal{L}^R_Q$, when Assumption \ref{ass:reparamTrick} is satisfied, one can use a popular reparametrization trick to derive a gradient for the critic parameters \(\theta\) as:
\begin{align}
\nabla_{\theta} & \mathcal{L}^R_Q(\theta)
= \nabla_{\theta} \mathbb{E}_{\substack{(s,a,r,s') \sim \mathcal{D} \\ \tilde{z} \sim F_z}} 
\left[ \bigl(\mathfrak{q}_\theta(s,a,\tilde{z}) - y(r,s')\bigr)^2 \right] \notag \\
&= \mathbb{E}_{\substack{(s,a,r,s') \sim \mathcal{D} \\ \tilde{z} \sim F_z}} 
\Bigl[ 2 \bigl(\mathfrak{q}_\theta(s,a,\tilde{z}) - y(r,s')\bigr) \notag   \nabla_{\theta} \mathfrak{q}_\theta(s,a,\tilde{z}) \Bigr]
\label{eq:critic_gradient_final_erick}
\end{align}

This gives rise to the stochastic update $\theta \;\leftarrow\; \theta - \eta_Q \, 2\big( \mathfrak{q}_\theta(s,a,\tilde{z}) - y(r,s') \big)\nabla_{\theta}\mathfrak{q}_\theta(s,a,\tilde{z})$. Optimizing \(\mathcal{J}^R_\pi\) is a bit more complex; we begin by letting \(q^*(s,\cdot;\phi)\) denote any statewise adversarial Q-value vector for policy \(\pi_\phi\):
\begin{equation}
\label{eq:qstar_def}
q^*(s,\cdot \,; \phi) \in \arg\min_{q \in \mathcal{U}_\theta(s)} \langle \pi_\phi(\cdot \mid s), q \rangle, \;\forall s\in\mathcal{S},
\end{equation}
which is well-defined due to compactness of $\mathcal{U}_\theta(s)$. Then, noting that the function 
\begin{align*}
f(\pi &) :=\;\mathbb{E}_{\substack{s \sim \mathcal{D} \\ a \sim \pi(\cdot \mid s)}} \bigg[ 
\min_{q \in \mathcal{U}_\theta(s)} \langle \pi(\cdot \mid s), q \rangle  - \alpha \log \pi(a \mid s) 
\bigg] \\
& =\hspace{-0.05cm}\mathbb{E}_{s \sim \mathcal{D}} \hspace{-0.05cm}\left[ 
\min_{q \in \mathcal{U}_\theta(s)} \langle \pi(\cdot\hspace{-0.05cm} \mid \hspace{-0.05cm}s), q \rangle  \hspace{-0.05cm}- \hspace{-0.05cm}\alpha \mathbb{E}_{a \sim \pi(\cdot \mid s)} \hspace{-0.05cm}\left[ \log \pi(a \hspace{-0.05cm}\mid \hspace{-0.05cm}s) 
\right]\right]
\end{align*}
is concave with respect to $\pi$, one can invoke the envelope theorem to identify one of its supergradients as 
\begin{align*}
\nabla_\pi \mathbb{E}_{s \sim \mathcal{D}} \biggl[
&\;\langle \pi(\cdot \mid s),\, q^*(s, \cdot\,; \phi) \rangle \\
&\;-\; \alpha\, \mathbb{E}_{a \sim \pi(\cdot \mid s)}
\bigl[\log \pi(a \mid s)\bigr]
\biggr]
\;\in\; \nabla_\pi f(\pi)
\end{align*}

We therefore obtain,  fixing $\bar{\phi}$ to $\phi$ that:

\begin{align}
\nabla_\phi \mathcal{J}_\pi^R(\phi)
= &\mathbb{E}_{s \sim \mathcal{D}} \biggl[
\sum_{a \in \mathcal{A}} q^*(s,a\,;\phi)\,
\nabla_\phi \pi_\phi(a \mid s) \notag
\\
&\quad - \alpha\,
\nabla_\phi \bigl\langle
\pi_\phi(\cdot \mid s),\,
\log \pi_\phi(\cdot \mid s)
\bigr\rangle
\biggr]
\label{eq:actor_gradient_final}
\end{align}
\removed{
We next expand the gradient term using the log-derivative trick. Since \( q^*(s,\cdot \,; \phi) \in \mathbb{R}^{|\mathcal{A}|} \) is fixed once the inner minimization is solved, the first gradient term in \eqref{eq:actor_gradient_final} can be written as:

\begin{equation}
\label{eq:score_term}
\begin{aligned}
\nabla_\phi \langle \pi_\phi(\cdot \mid s)&, q^*(s,\cdot \,; \phi) \rangle = \mathbb{E}_{a \sim \pi_\phi(\cdot \mid s)} 
\left[ q^*(s,a; \phi) \nabla_\phi \log \pi_\phi(a \mid s)  \right]
\end{aligned}
\end{equation}

Putting everything together, we obtain the robust actor gradient as,
\begin{equation}
\label{eq:actor_gradient_final_expanded}
\begin{aligned}
\nabla_\phi J(\phi) = \mathbb{E}_{s \sim \mathcal{D}} \big[
&\mathbb{E}_{a \sim \pi_\phi(\cdot \mid s)} \big[
\nabla_\phi \log \pi_\phi(a \mid s) \cdot \big(
q^*(s,a; \phi)  \big] \\
&- \alpha \, \mathbb{E}_{a \sim \pi_\phi(\cdot \mid s)} 
\left[ \nabla_\phi \log \pi_\phi(a \mid s) \right] \big]
\end{aligned}
\end{equation}
\EDcomments{I get:
\begin{equation}
\label{eq:actor_gradient_final_expanded:ED}
\begin{aligned}
\nabla_\phi J(\phi) = \mathbb{E}_{s \sim \mathcal{D}} \big[
&\mathbb{E}_{a \sim \pi_\phi(\cdot \mid s)} \big[
\nabla_\phi \log \pi_\phi(a \mid s) \cdot \big(
q^*(s,a; \phi)  \big] \\
&- \alpha \, \mathbb{E}_{a \sim \pi_\phi(\cdot \mid s)} 
\left[ (1+\log\pi_\phi(a \mid s)) \nabla_\phi \log \pi_\phi(a \mid s) \right] \big]
\end{aligned}
\end{equation}
}
}
This produces a standard entropy-regularized policy gradient, but is evaluated with respect to the worst-case value vector \( q^*(s,\cdot \,; \phi) \) in the uncertainty set, providing robustness to epistemic uncertainty. We summarize the training procedure for Robust SAC-N in Algorithm~\ref{alg:robust_sacn} in Appendix~\ref{sec:appendix_algorithms}.

\section{Sample-based construction of $\mathcal{U}_\theta(s)$ from $\mathfrak{q}_\theta(s,a,\tilde{z})$}
\label{sec:set_geometry_conservativeness}

In practice, one often approximates $F_\theta^q(s)$ using Monte Carlo samples, which form an empirical distribution $\widehat{F}_\theta^q(s)$. 
Having access to  $\widehat{F}_\theta^q(s)$, one can  approximate $\mathcal{U}(F_\theta^q(s))$ with $\mathcal{U}(\widehat{F}_\theta^q(s))$. 
Different choices of \( \mathcal{U}(\widehat{F}^q_\theta(s)) \) lead to varying trade-offs between computational tractability, policy sensitivity, and expressiveness. In the remainder of this section, we present three popular sets from the literature of robust optimization: box, convex hull and ellipsoidal sets. 

\textbf{Box set:} Let $\{\tilde{z}_i\}_{i=1}^N$ be $N$ values sampled from $F_z$. The simplest construction is the box set introduced in \eqref{eq:box_set_def}, which defines $\mathcal{U}_\theta(s)$ as the Cartesian product of the intervals covering $\tilde{q}(a)$ for each action. In a sample-based setting, this reduces to :
\begin{equation}
\label{eq:box_set}
\begin{aligned}
\mathcal{U}_{\text{box}}(\widehat{F}_\theta^q(s)) :=
\mathop{\times}_{a \in \mathcal{A}} \biggl[\min_{i} \mathfrak{q}_\theta(s,a,\tilde{z}_i), \max_{i} \mathfrak{q}_\theta(s,a,\tilde{z}_i)
\biggr]
\end{aligned}
\end{equation}


\textbf{Convex Hull Set:}
A more expressive alternative is the uncertainty set operator that produces the convex hull of the support of $F_\theta^q(s)$. In a sample-based setting, this reduces to:
\begin{equation}
\label{eq:ch_set}
\begin{aligned}
\mathcal{U}_{\text{hull}}\!\left(\widehat{F}_\theta^q(s)\right)&
:= {} 
\biggl\{
\sum_{i=1}^N \lambda_i\, \mathfrak{q}_\theta(s,\cdot,\tilde{z}_i)
\\
& \;\biggm|\;
\exists\, \lambda \in \mathbb{R}^N,\ 
\lambda_i \ge 0\ \forall i,\ 
\sum_{i=1}^N \lambda_i = 1
\biggr\}
\end{aligned}
\end{equation}

The worst-case Q-vector is \( q^*(s,a;\phi) = \mathfrak{q}_\theta(s,a,z^*(s,\phi)) \), where \( z^*(s,\phi) \in \arg\min_i \mathbb{E}_{a \sim \pi_\phi(\cdot \mid s)}[\mathfrak{q}_\theta(s,a,\tilde{z}_i)] \).

\textbf{Ellipsoidal Set:}
In this work, we will mainly consider an ellipsoidal set operator that aim to cover a certain proportion $\upsilon$ of the total mass of $F_\theta^q(s)$. In a sample-based setting, this can be done by estimating the empirical mean and covariance of the sampled Q-vectors:
\begin{equation}
\begin{aligned}
\hat{\mu}(s)
&:= \tfrac{1}{N} \sum_{i=1}^N \mathfrak{q}_\theta\!\left(s,\cdot,\tilde{z}_i\right), \\[0.3em]
\widehat{\Sigma}(s)
&:= \tfrac{1}{N} \sum_{i=1}^N
\Bigl(
\mathfrak{q}_\theta\!\left(s,\cdot,\tilde{z}_i\right) - \hat{\mu}(s)
\Bigr)
\Bigl(
\mathfrak{q}_\theta\!\left(s,\cdot,\tilde{z}_i\right) - \hat{\mu}(s)
\Bigr)^{\!\top}.
\end{aligned}
\end{equation}

and estimating the radius as 

\[
\begin{aligned}
\widehat{\Upsilon}(s)
:= \inf \Biggl\{ \Upsilon \;\Bigg|\;
&\frac{1}{N} \sum_{i=1}^N \mathbf{1} \Biggl\{ \bigl( \mathfrak{q}_\theta(s,\cdot,\tilde{z}_i) - \hat{\mu}(s) \bigr)^\top
\widehat{\Sigma}(s)^{-1} \\
& \cdot
\bigl( \mathfrak{q}_\theta(s,\cdot,\tilde{z}_i) - \hat{\mu}(s) \bigr)
\leq \Upsilon^2
\Biggr\}
\geq \upsilon
\Biggr\}.
\end{aligned}
\]
The corresponding uncertainty set is defined as:
\begin{equation}
\label{eq:ell_set}
\begin{aligned}
\mathcal{U}_{\text{ell}}\!\left(\widehat{F}_\theta^q(s)\right)
:= \biggl\{\, q \in \mathbb{R}^{|\mathcal{A}|} \,\bigg|\,
& (q - \hat{\mu}(s))^\top\, \widehat{\Sigma}(s)^{-1} \\
&  \cdot (q - \hat{\mu}(s))
\;\le\; \widehat{\Upsilon}(s)^2
\biggr\}
\end{aligned}
\end{equation}
This set encodes second-order structure and supports efficient optimization. When $\widehat{\Sigma}(s)$ is positive definite, the worst-case Q-vector under a given policy admits the closed-form solution:
\[
q^*(s,\cdot\,; \phi) = \hat{\mu}(s) - \widehat{\Upsilon}(s)\cdot \frac{\widehat{\Sigma}(s)\pi_\phi(\cdot \mid s)}{\|\widehat{\Sigma}(s)^{1/2} \pi_\phi(\cdot \mid s)\|}.
\]

For completeness, the detailed derivations of the policy-sensitive worst-case Q-vector under both the convex hull and ellipsoidal sets are provided in Appendix~\ref{appendix:q_star_derivations}.

We refer the reader to Appendix~\ref{sec:appendix_algorithms} for the pseudocode of the training algorithm based on box, convex hull (Algorithm~\ref{alg:ERSAC-box-hull}) and ellipsoidal (Algorithm~\ref{alg:ERSAC-E}) uncertainty sets.
A deeper discussion on how the choice of uncertainty set affects the sensitivity of the worst-case Q-vector to the policy \(\pi_\phi\), based on the Machine Replacement example introduced earlier, is provided in Appendix~\ref{appendix:set_comparision}.

\section{The ERSAC model with Epinet (ERSAC(Epi)) }
\label{sec:ERSAC-Epi}

Recall from Assumption~\ref{ass:reparamTrick} that we require a parametric sampling operator \( \mathfrak{q}_\theta(s,a, z) \), with \( z \sim F_z \), such that $\mathfrak{q}_\theta(s, \cdot, z) \sim F^q_\theta(s),$ where \( F^q_\theta(s) \in \mathcal{M}(\mathbb{R}^{|\mathcal{A}|}) \) denotes a distribution over Q-value vectors. We instantiate this generative model using an Epistemic Neural Network (Epinet) introduced by \cite{osband2023epistemic}, which enables structured and differentiable sampling from a single neural network. An Epinet supplements a base network \( \mu_{\theta_\mu}(s,a) \in \mathbb{R}\), parameterized by \( \theta_\mu \), which yields the mean Q-value vector. From this base, we extract a feature representation \( \psi_{\theta_\mu}(s) \in \mathbb{R}^{d_\psi} \), typically taken from the last hidden layer. Epistemic variation is introduced via a latent index \( z \sim \mathcal{N}(0, I) \in \mathbb{R}^{d_z} \). These components are combined through a stochastic head \( \sigma_{\theta_\sigma}(\psi_{\theta_\mu}(s), a,z) \in \mathbb{R} \), 
which modulates the structured uncertainty. The sampling operator for the Q-value vector is then defined as $\mathfrak{q}_\theta(s, \cdot, z) := \mu_{\theta_\mu}(s,\cdot) + \sigma_{\theta_\sigma}(\psi_{\theta_\mu}(s), \cdot, z), \label{eq:epinet_q_sample} $. The stochastic head is constructed as $\sigma_{\theta_\sigma}(\psi, \cdot, z) := \sigma^\text{L}_{\theta_\sigma}(\psi,\cdot,z) + \sigma^\text{P}(\psi,\cdot,z) \label{eq:epinet_sigma}$ with \( \sigma^\text{L}_{\theta_\sigma}: \mathbb{R}^{d_\psi}\times \mathcal{A}\times \mathbb{R}^{d_z} \to \mathbb{R} \) as a learnable function and \( \sigma^\text{P}: \mathbb{R}^{d_\psi}\times\mathcal{A}\times \mathbb{R}^{d_z} \to \mathbb{R} \) as a fixed prior. The fixed prior network \( \sigma^\text{P} \) encodes initial epistemic uncertainty by inducing variability in predictions across samples of indices \( z \).  In well explored regions, \( \sigma^\text{L}_{\theta_\sigma} \) can learn better distributions for the predictive uncertainty, while in data sparse areas, \( \sigma^\text{P} \) can induce the prior beliefs of the decision maker to guide conservative predictions. We can now use it to generate the realizations of the Q-value vectors at a given state $s$ by drawing $z \sim \mathcal{N}(0,I)$ to form the empirical distribution $\widehat{F}_{\theta}(s)$ over Q values. This enables us to employ the sample based epistemic uncertainty sets introduced in the earlier section.

\removed{
\EDcomments{Later we can study the special case where:
\begin{equation}
\sigma^\text{L}_\theta(\psi_\theta(s),z) := A^L_\theta(\psi_\theta(s))^Tz \text{ and } \sigma^\text{P}(\psi_\theta(s),z):= A^P(\psi_\theta(s))^Tz
\end{equation}
for some neural networks \( A^\text{L}_\eta: \mathbb{R}^{d_\psi} \to \mathbb{R}^{d_z \times |\mathcal{A}|} \) and \( A^\text{P}: \mathbb{R}^{d_\psi} \to \mathbb{R}^{d_z \times |\mathcal{A}|} \).
}

\EDcomments{Used the following for the special case:
Together, they define uncertainty sets that are compact where the model is confident and expansive where uncertainty remains high. This formulation induces a Gaussian distribution,
\begin{equation}
\mathfrak{q}_\theta(s, z, c) \sim \mathcal{N}(\mu_\zeta(s),\, A(s)^\top  A(s)), \label{eq:epinet_distribution}
\end{equation}
providing a simple and expressive mechanism for sampling Q-value vectors. By drawing independent samples \( (z, c) \sim \mathcal{N}(0, I) \times \text{Unif}(\mathbb{S}^{d_z})^{|\mathcal{A}|} \), we obtain diverse predictions that reflect epistemic variability without resorting to an explicit ensemble. }}

This construction yields a parameter efficient and fully differentiable reparameterization of the Q distribution. Further, one can train these networks using a perturbed squared loss inspired by Gaussian bootstrapping following the loss:
\begin{equation}
\label{eq:enn_loss}
\begin{aligned}
\mathcal{L}_Q^{\text{ENN}}(\theta)
:=\;&
\mathbb{E}_{(s,a,r,s',c)\sim\bar{\mathcal{D}},\,\tilde{z}\sim F_z}
\Bigl[
\bigl(
\mathfrak{q}_\theta(s,a,\tilde{z})
- y(r,s') \\
&- \bar{\sigma}\,\langle c,\tilde{z}\rangle
\bigr)^2
\Bigr]
+ \lambda_\mu \,\|\theta_\mu\|^2
+ \lambda_\sigma \,\|\theta_\sigma\|^2 .
\end{aligned}
\end{equation}

where each member $(s,a,r,s')$ from the dataset $\mathcal{D}$ is augmented with some $c$ randomly sampled from the surface of the unit sphere \( \mathbb{S}^{d_z} \) to produce $\bar{\mathcal{D}}$, 
where \( \bar{\sigma} > 0 \) denotes the bootstrap noise scale, and where \( \lambda_\zeta, \lambda_\eta \) are regularization coefficients. This loss encourages the network to match bootstrapped Q-targets while introducing variability across $z$ samples. It can be minimized via standard stochastic gradient methods. The ENN critic updates thus become:
\begin{align}
\theta_\mu \leftarrow\ &\theta_\mu
\;-\;2\eta_Q \cdot
\biggl(
\tfrac{1}{|\mathcal{B}|}
\sum_{(s,a,r,s',c) \in \bar{\mathcal{B}}}
\mathbb{E}_{\tilde{z} \sim F_z}
\Bigl[
\mathfrak{q}_\theta(s,a,\tilde{z}) \nonumber \\
&- y(r,s') - \bar{\sigma} \langle c,\tilde{z} \rangle
\Bigr]
\cdot \nabla_{\theta_\mu} \mu_{\theta_\mu}(s,a)
\biggr)\;-\;4\eta_Q \lambda_\mu \theta_\mu
\label{eq:mu_update}
\end{align}

\begin{align}
&\theta_\sigma \leftarrow\ \theta_\sigma
\;-\;2\eta_Q \cdot
\biggl(
\tfrac{1}{|\mathcal{B}|}
\sum_{(s,a,r,s',c) \in \bar{\mathcal{B}}}
\mathbb{E}_{\tilde{z} \sim F_z}
\Bigl[
\mathfrak{q}_\theta(s,a,\tilde{z}) \nonumber \\
&- y(r,s') - \bar{\sigma} \langle c,\tilde{z} \rangle
\Bigr]
\cdot \nabla_{\theta_\sigma}
\sigma^L_{\theta_\sigma}\bigl(\psi_{\theta_\mu}(s),a,\tilde{z}\bigr)
\biggr) - 4\eta_Q \lambda_\sigma \theta_\sigma
\label{eq:sigma_update}
\end{align}

To accelerate the evaluation of $\mathcal{U}(F_\theta^q(s)$ when using an ellipsoidal uncertainty set operator, we introduce additional structure in $\sigma^\text{L}_{\theta_\sigma}(\psi,\cdot,z)$ and $\sigma^\text{P}(\psi,\cdot,z)$ as outlined in Assumption \ref{ass:linearENN}, namely that both operators are linear in $z$.
\begin{assumption}\label{ass:linearENN}
The stochastic heads $\sigma^\text{L}_{\theta_\sigma}(\psi, a, z)$ and $\sigma^\text{P}(\psi, a, z)$ are linear in $z$, i.e.,
\[
\sigma^\text{L}_{\theta_\sigma}(\psi, a, z) = \langle \bar{\sigma}^\text{L}_{\theta_\sigma}(\psi, a), z \rangle, \
\sigma^\text{P}(\psi, a, z) = \langle \bar{\sigma}^\text{P}(\psi, a), z \rangle,
\]
for some mappings $\bar{\sigma}^\text{L}_{\theta_\sigma} : \mathbb{R}^{d_\psi} \times \mathcal{A} \to \mathbb{R}^{d_z}$ and $\bar{\sigma}^\text{P} : \mathbb{R}^{d_\psi} \times \mathcal{A} \to \mathbb{R}^{d_z}$.
\end{assumption}
Assumption \ref{ass:linearENN} induces a Gaussian distribution,
\begin{equation}
\mathfrak{q}_\theta(s, \cdot, z) \sim \mathcal{N}(\mu_{\theta_\mu}(s),\, \Sigma_{\theta}(s)), \label{eq:epinet_distribution}
\end{equation}
where the covariance is defined as, $\left[\Sigma_{\theta}(s)\right]_{a,a'}:=\langle \bar{\sigma}^L_{\theta_\sigma}(\psi_{\theta_\mu}(s), a)+\bar{\sigma}^P(\psi_{\theta_\mu}(s), a), \bar{\sigma}^L_{\theta_\sigma}(\psi_{\theta_\mu}(s), a')+\bar{\sigma}^P(\psi_{\theta_\mu}(s), a')\rangle$. This gives rise to the Epinet based ellipsoidal set:
\begin{align}
\mathcal{U}_{\text{ell}}^{\text{ENN}}(s)
:= \Biggl\{
q \in &\mathbb{R}^{|\mathcal{A}|} \,\bigg|\;
\bigl(q - \mu_{\theta_\mu}(s)\bigr)^{\!\top}
\,\Sigma_{\theta}(s)^{-1} \nonumber \\[0.3em]
&\cdot \bigl(q - \mu_{\theta_\mu}(s)\bigr)\le F^{-1}_{\chi^2_{|\mathcal{A}|}}(\upsilon)
\Biggr\}
\label{eq:epinet_ellipsoid}
\end{align}
Here, \(F_{\chi^2_{|\mathcal{A}|}}^{-1}(\upsilon)\) denotes the inverse CDF of the \(\chi^2\) distribution with \(|\mathcal{A}|\) degrees of freedom, yielding an efficient alternative to ensemble based uncertainty modeling with a closed form worst case Q-vector. The assumption of linear stochastic heads in Epinet is mainly for computational efficiency, allowing closed-form mean and covariance estimates for ellipsoidal uncertainty sets. While this may limit expressivity compared to nonlinear heads, it is generally sufficient for capturing epistemic uncertainty in many RL settings. In highly non-Gaussian cases, richer parameterizations or sampling-based approaches may be needed. Relaxing this assumption could enable more flexible uncertainty modeling, but at increased computational cost.

The training procedure for ERSAC with Epinet (ERSAC(Epi)) mirrors the ensemble based variant (Algorithm~\ref{alg:ERSAC-E}) but avoids sampling by leveraging the structured Epinet model. The mean and covariance are directly obtained as \( \mu_{\theta_\mu}(s) \) and \( \Sigma_\theta(s) \) from the deterministic and stochastic heads under Assumption~\ref{ass:linearENN}. The ellipsoidal radius is set to \( \Upsilon^2(s) = F^{-1}_{\chi^2_{|\mathcal{A}|}}(\upsilon) \), ensuring a \(\upsilon\)-level confidence set. This enables efficient, fully differentiable updates for both the Bellman target and policy gradient. See Appendix~\ref{sec:appendix_algorithms}, Algorithm~\ref{alg:ERSAC-E-Epi} for full details.

\section{Experiments}
\label{sec:experiments}
This section presents a comprehensive empirical evaluation of our framework for epistemic robustness in offline reinforcement learning. Epistemic uncertainty is captured via uncertainty sets that integrate seamlessly into robust policy optimization. The three sample-based uncertainty sets 
lead to three ERSAC variants: \textbf{SAC-N} 
(ERSAC with a box set over \(N\) ensembles), \textbf{ERSAC-CH-N} (convex hull over ensembles), and \textbf{ERSAC-Ell-N} (ellipsoids from empirical mean and covariance). We also evaluate \textbf{ERSAC-Ell-Epi}, 
which replaces the ensemble with $N$ samples from \textbf{ERSAC-Ell-N} to produce a sample-based ellipsoid. 
Lastly, \textbf{ERSAC-Ell-Epi*} leverages the structured stochastic head \( \sigma_{\theta_\sigma}(\psi, \cdot, z) \) (see Assumption~\ref{ass:linearENN}) to construct ellipsoidal sets directly, without sampling. The code can be found on GitHub\footnote{\url{https://github.com/Achenred/ERSAC}}.

Our experiments span a diverse set of environments, including tabular domains (Machine Replacement and Riverswim), classic control benchmarks (CartPole and LunarLander) and Atari environments. Across these domains, we evaluate each method's ability to learn effective policies under distributional shifts arising due to changes in behavior policies and limited data coverage.

A key contribution of our work is a novel offline RL benchmarking framework that enables control over the risk sensitivity of the behavior policy used to generate offline datasets. By adjusting the level of optimism or pessimism through expectile-based value learning, we can systematically evaluate how the nature of behavioral data affects the performance of offline RL algorithms.
To induce risk sensitivity, we employ a modified actor-critic algorithm incorporating the dynamic expectile risk measure (\citet{marzban2023deep}). 
For each \((s, a)\), critic target is computed using a bootstrapped expectile estimate,
\vspace{-0.5em}
\[
\begin{aligned}
y := \arg\min_{z \in \mathbb{R}}\;
\sum_{j=1}^M
\Big|
\mathbb{I}\!\big(
z < r(s,a) + \gamma \max_{a'} Q_\theta(s_j',a')
\big) - \tau
\Big| \\
\smash{\cdot}\;
\Big(
z - r(s,a) - \gamma \max_{a'} Q_\theta(s_j',a')
\Big)^2
\end{aligned}
\]
\vspace{-0.5em}
and the critic minimizes squared error to this target. 
The actor is trained via a standard policy gradient to maximize expected Q-values. After a fixed number of training steps, the resulting policy \(\pi_\phi\) reflects the desired level of risk sensitivity through \(\tau\). We then collect an offline dataset of size \(N\) using \(\varepsilon\)-greedy interaction with the environment, selecting random actions with probability \(\varepsilon = 0.1\). This yields datasets with systematically varying behavioral bias. Full implementation details are provided in Appendix~\ref{alg:expectile_data_gen}.

\subsection{Evaluation on tabular tasks}

We first evaluate ERSAC on two tabular MDPs, \textit{Machine Replacement} and \textit{Riverswim}, which provide interpretable structure while capturing core offline RL challenges such as sparse state–action coverage and sensitivity to policy extrapolation. The tabular setting isolates epistemic uncertainty without confounding deep RL effects, enabling a clean comparison of uncertainty set constructions.

Offline datasets are generated by varying (i) dataset size ($10, 100, 1000 \times |\mathcal{S}|$) and (ii) behavior policy risk sensitivity using dynamic expectiles $\tau \in \{0.1, 0.5, 0.9\}$, inducing systematic differences in coverage. Performance is measured using normalized returns, computed relative to a random and optimal policy, averaged over 100 evaluation episodes.

Table~\ref{tab:aggregated_returns} reports normalized returns aggregated over $\tau$ for each dataset size. In low-data regimes, structured uncertainty sets (CH-N, Ell$_{0.9}$-N) outperform the box baseline (B-N) by up to $75\%$, highlighting the importance of modeling epistemic structure under sparse coverage. As dataset size increases, all methods improve, but convex and ellipsoidal sets converge faster to optimal performance.

Under risk-averse behavior policies ($\tau=0.9$), where epistemic uncertainty is highest, ellipsoidal variants remain robust. Comparing ellipsoids covering $100\%$ versus $90\%$ of ensemble samples, the tighter Ell$_{0.9}$-N consistently performs better, likely by filtering outlier critics and avoiding over-pessimism. We therefore adopt $90\%$ coverage in subsequent experiments.

\begin{table*}[t]
  \centering

  \renewcommand{\arraystretch}{1.15}
\scalebox{0.8}{
  \begin{subtable}[t]{\textwidth}
    \centering
    \begin{tabular}{l|c|rrrr|r}
      \hline
      \textbf{Env} & \textbf{DS} &
      \textbf{SAC-N} & \textbf{CH-N} & \textbf{Ell-N} & \textbf{Ell\_0.9-N} & \textbf{Beh. Policy} \\
      \hline
      \multirow{3}{*}{\textbf{Machine Replacement}} 
        & 10$\times$   & \underline{$84 \pm 3$} & $86 \pm 2$ & $89 \pm 2$ & $\mathbf{90 \pm 2}$ & $93 \pm 2$ \\
        & 100$\times$  & $97 \pm 2$ & $\mathbf{96 \pm 2}$ & \underline{$94 \pm 2$} & $95 \pm 2$ & $93 \pm 2$ \\
        & 1000$\times$ & $97 \pm 2$ & $97 \pm 2$ & \underline{$97 \pm 2$} & $\mathbf{97 \pm 1}$ & $93 \pm 2$ \\
      \hline
      \multirow{3}{*}{\textbf{RiverSwim}} 
        & 10$\times$   & \underline{$47 \pm 3$} & $58 \pm 3$ & $55 \pm 3$ & $\mathbf{60 \pm 3}$ & $5 \pm 4$ \\
        & 100$\times$  & \underline{$96 \pm 2$} & $97 \pm 2$ & $97 \pm 2$ & $\mathbf{98 \pm 2}$ & $5 \pm 4$ \\
        & 1000$\times$ & $99 \pm 1$ & $99 \pm 1$ & $100 \pm 0$ & $\mathbf{100 \pm 0}$ & $5 \pm 4$ \\
      \hline
    \end{tabular}
    \caption{Tabular environments}
    \label{tab:aggregated_returns}
  \end{subtable}
}
  \vspace{1em}
\scalebox{0.8}{
  \begin{subtable}[t]{\textwidth}
    \centering
    \begin{tabular}{l|c|rrrrr|r}
      \hline
      \textbf{Env} & \textbf{DS} &
      \textbf{SAC-N} & \textbf{CH-N} & \textbf{Ell\_0.9-N} & \textbf{Ell-Epi} & \textbf{Ell-Epi$^*$} & \textbf{Beh. Policy} \\
      \hline
      \multirow{3}{*}{\textbf{CartPole}} 
        & 1k   & \underline{$76 \pm 3$} & $74 \pm 2$ & $\mathbf{79 \pm 2}$ & $79 \pm 2$ & $77 \pm 2$ & $90 \pm 2$ \\
        & 10k  & \underline{$96 \pm 2$} & $98 \pm 1$ & $\mathbf{100 \pm 0}$ & $\mathbf{100 \pm 0}$ & $\mathbf{100 \pm 0}$ & $90 \pm 2$ \\
        & 100k & $\mathbf{100 \pm 0}$ & $\mathbf{100 \pm 0}$ & $\mathbf{100 \pm 0}$ & $\mathbf{100 \pm 0}$ & $\mathbf{100 \pm 0}$ & $90 \pm 2$ \\
      \hline
      \multirow{3}{*}{\textbf{LunarLander}} 
        & 1k   & \underline{$69 \pm 2$} & $74 \pm 2$ & $97 \pm 2$ & $97 \pm 2$ & $\mathbf{97 \pm 2}$ & $89 \pm 3$ \\
        & 10k  & \underline{$93 \pm 2$} & $99 \pm 2$ & $101 \pm 1$ & $100 \pm 2$ & $\mathbf{102 \pm 1}$ & $89 \pm 3$ \\
        & 100k & \underline{$98 \pm 2$} & $100 \pm 2$ & $104 \pm 1$ & $\mathbf{107 \pm 2}$ & $106 \pm 1$ & $89 \pm 3$ \\
      \hline
    \end{tabular}
    \caption{Gym environments}
    \label{tab:aggregated_gym}
  \end{subtable}
  }

  \caption{Returns aggregated across $\tau \in \{0.1, 0.5, 0.9\}$ for each dataset size. Bold indicates best method, underline the worst, when mean differences $\geq 1$.}
  \label{tab:aggregated_combined}
\end{table*}

\vspace{-0.7em}
\subsection{Evaluation on Gym environments}

We next evaluate the proposed methods on two widely used Gym environments, \textit{CartPole} and \textit{LunarLander}. CartPole is a standard control task with binary rewards and continuous states, while LunarLander presents greater complexity with shaped rewards and a higher-dimensional state-action space. As in the tabular setting, we construct offline datasets by varying two factors: dataset size and behavior policy risk profile. For each environment, we generate nine datasets by crossing three dataset sizes (1K, 10K, and 100K transitions) with three expectile levels: \(\tau = 0.1\) (risk-seeking), \(\tau = 0.5\) (risk-neutral), and \(\tau = 0.9\) (risk-averse). Behavior policies are trained to convergence using a dynamic expectile based actor-critic algorithm, and fixed trajectories are collected for each configuration.

Table~\ref{tab:aggregated_gym} summarizes normalized returns aggregated over $\tau$ values for each dataset size, while full results across all $\tau$ settings are provided in Table~\ref{tab:results_gym} in Appendix~\ref{appendix:results}. We consider the policy trained under the risk neutral behavior($\tau = 0.5$) as the reference optimal policy. First, models CH-N, Ell\_0.9-N, Ell-Epi  consistently outperform the box baseline B-N, particularly in data scarce and risk averse settings where epistemic uncertainty plays a larger role. When we aggregate returns across dataset sizes by risk level (As presented in Table~\ref{tab:ell09n_agg}), we observe that Ell\_0.9-N consistently achieves strong performance under risk-neutral and risk-seeking behavior policies, suggesting that the method effectively leverages optimistic data to enhance policy learning.

\begin{table}[h]
\centering

\scalebox{0.9}{%
\begin{tabular}{lccc}
\toprule
\textbf{Env} & $\tau{=}0.1$ & $\tau{=}0.5$ & $\tau{=}0.9$ \\
\midrule
CartPole    & $95\pm8$ (1)  & $93\pm14$ (2) & $92\pm14$ (3) \\
LunarLander & $103\pm7$ (1) & $99\pm5$ (2)  & $99\pm5$ (2) \\
MR          & $93\pm1$ (3)  & $95\pm4$ (1)  & $94\pm3$ (2) \\
RS          & $87\pm15$ (2) & $87\pm17$ (1) & $84\pm22$ (3) \\
\bottomrule
\end{tabular}
}
\caption{Agg. performance of \textbf{Ell\_0.9-N} across environments with mean $\pm$ std and within-environment rank (1 = best).}
\label{tab:ell09n_agg}
\end{table}
Ellipsoidal variants show strong, often best, performance across settings. Ell-Epi$^{*}$ matches or outperforms the ensemble based Ell\_0.9-N in several cases, highlighting Epinet-based uncertainty as an efficient alternative. We observed that Ell-Epi$^{*}$ achieves comparable performance with significantly lower compute (see Appendix \ref{appendix:results} for details), making it attractive for scaling to complex domains.

To further understand how uncertainty sets affect learning dynamics, we analyze policy entropy during training. We observed that Box-based methods (B-N) maintain consistently lower entropy, indicating less stochastic and more prematurely deterministic policies. This often leads to suboptimal convergence. In contrast, CH-N, Ell-N, and Ell-Epi allow more flexible shaping of \(q^*(s, \cdot\,; \phi)\), encouraging exploration and enabling better identification of high-reward actions under offline constraints. We refer the reader to Appendix \ref{appendix:results} for a detailed report.

\subsection{Evaluation on Atari environments}
To assess scalability beyond tabular and classic-control settings, we additionally evaluate ERSAC on five Atari~2600 environments.
The goal here is to evaluate the scalability of ERSAC's epistemic robust value estimation to high dimensional domains and noisy, heterogeneous data typically found in Atari offline datasets. We use standard Atari offline datasets sourced from Minari~\cite{minari}, which provide trajectories collected from diverse and partially suboptimal behavior policies.

These experiments highlight the advantages of ERSAC models in diverse settings. The Ell-Epi$^*$ variant achieves the strongest scores in Seaquest and Hero environments, suggesting that it handles over estimation of Q values more effectively in ambiguous environments where reward sparsity and bootstrapping noise amplify estimation risk. In more predictable games such as Pong and Breakout, Ell-Epi$^*$ matches the performance of CQL and IQL, indicating that its uncertainty sets naturally contract when epistemic uncertainty is low and avoid the excessive pessimism that can hinder conservative methods. In Q$^{*}$bert, where long horizon return propagation and irregular rewards create substantial uncertainty, the ERSAC models close much of the gap to IQL, demonstrating the benefit of structured uncertainty modeling over other baselines. Across all environments, ERSAC models consistently outperforms BRAC-BCQ, and notably, Ell-Epi$^*$ ranks within the top three methods in all games, reflecting more reliable handling of unsupported state action pairs and high variance value targets.

Overall, the results show that Ell-Epi$^*$ scale effectively to high dimensional domains, reinforcing structured epistemic modeling as a principled foundation for offline RL in complex environments. Full experimental details and results are deferred to Appendix \ref{appendix:atari}.

\section{Conclusion}

We introduce \textit{Epistemic Robust Soft Actor-Critic} (ERSAC), a unified offline reinforcement learning framework that models epistemic uncertainty via uncertainty sets over $Q$-values, replacing ensemble-based pessimism with structured box, convex hull, and ellipsoidal constructions. ERSAC enables conservative yet flexible value estimation and policy optimization, generalizing SAC-N as a special case while exposing trade-offs between expressiveness and computational cost across set geometries. An Epinet-based variant yields closed-form ellipsoidal uncertainty sets, significantly reducing runtime without sacrificing performance. 

By leveraging risk-aware behavior policies, ERSAC systematically induces coverage bias in offline datasets, allowing controlled modulation of epistemic uncertainty and conservative value estimation. Empirically, ERSAC uncertainty sets are most effective under poor or biased coverage, with uncertainty shrinking as data coverage improves, at which point performance approaches that of standard ensemble methods. Beyond benchmarking robustness in offline RL, this framework offers a foundation for studying epistemic robustness under risk-sensitive behavior policies. Promising future directions include extending ERSAC to multi-agent and hierarchical reinforcement learning, incorporating risk-aware objectives, and establishing finite-sample generalization guarantees and regret bounds under epistemic uncertainty. Overall, ERSAC demonstrates that structured and efficient epistemic modeling is a viable path toward safe, generalizable, and scalable offline reinforcement learning.

\newpage



\bibliography{references}
\bibliographystyle{icml2026}

\newpage
\appendix
\onecolumn

\section{Appendix}
This appendix provides literature context, theoretical foundations, algorithmic details, and extended empirical results that support our main findings.

We begin in Section~\ref{appendix:related-work} with a review of related work on epistemic uncertainty modeling and robust offline reinforcement learning. Section~\ref{appendix:replacement-example} analyzes the state visitation frequencies in the Machine Replacement problem under various behavior policies introduced in the main text. We further build on this example to study the sensitivity of the worst-case Q-function to the policy \(\pi_\phi\).

Section~\ref{appendix:sacn-special-case} presents a formal lemma and proof showing that SAC-N is a special case of our proposed framework. Section~\ref{appendix:q_star_derivations} derives closed-form expressions for the worst-case Q-vectors induced by convex hull and ellipsoidal sets.

Section~\ref{sec:appendix_algorithms} provides pseudocode for the ERSAC algorithmic variants proposed in this work. Section~\ref{appendix:data_generation} describes the offline data generation process under different behavior policies. Section~\ref{appendix:training_details} details the experimental setup, including training procedures and hyperparameters. Finally, Section~\ref{appendix:results} presents full empirical results across environments, dataset sizes, and risk sensitivity levels, complementing the main text with additional tables and figures.

\subsection{Literature review}
\label{appendix:related-work}
While the \textbf{motivation for offline RL} originates primarily from safety, cost, and deployment constraints in domains such as healthcare, robotics, and industrial control, recent work highlights its broader benefits, including improved generalization and sample efficiency when combined with online learning \cite{ball2023efficient, jelley2024efficient}. Offline data can stabilize learning and accelerate convergence through pretraining or regularization \cite{kumar2022should}. However, the absence of environment interaction exacerbates challenges like overestimation and error compounding, especially when using deep value function approximators. These failures are often attributed to epistemic uncertainty in out of distribution state-action pairs, where neural networks are known to make overconfident predictions \cite{lakshminarayanan2017simple, kendall2017uncertainties}. Ensemble-based and Bayesian methods partially mitigate this by explicitly modeling uncertainty, highlighting the need for structured epistemic reasoning in offline settings. 

\textbf{Model-free methods} primarily focus on constraining the learned policy or value estimates to remain within the support of the dataset, thereby mitigating extrapolation errors. One class of such methods, known as policy constraint methods, restricts the learned policy to stay close to the behavior policy. This reduces the likelihood of selecting actions not well represented in the data. Approaches like BCQ \cite{fujimoto2018addressing}, BEAR \cite{kumar2019stabilizing}, and BRAC \cite{wu2019behavior} explicitly enforce such constraints using divergence penalties or support matching. Another class focuses on value regularization, where conservative value estimates discourage overoptimistic Q-values for out-of-distribution actions. Notably, CQL \cite{kumar2020conservative} enforces a soft lower-bound on Q-values, while EDAC \cite{an2021uncertainty} and other ensemble-based methods use Q-function diversity to reduce overestimation risk. More recent work has revisited how generalization influences error propagation in offline RL.
DMG~\cite{mao2024doubly} shows that limited extrapolation beyond the dataset can be beneficial when properly controlled, introducing a doubly‑mild Bellman backup that blends in‑sample and mildly generalized actions to reduce overestimation without fully suppressing generalization.
A closely related line of work targets distribution shift in both states and actions.
SCAS~\cite{mao2024offline} performs OOD state correction using learned dynamics while simultaneously suppressing OOD actions, offering a unified mechanism for preventing harmful extrapolation during policy improvement.

\textbf{Model-based methods} instead aim to learn an explicit model of the environment's dynamics, which can be used for policy learning or evaluation via simulated rollouts. Examples include MOPO \cite{yu2020mopo}, which penalizes uncertainty in model rollouts, and MOReL \cite{kidambi2020morel}, which builds a pessimistic MDP based on model confidence. COMBO \cite{yu2021combo} combines model-based rollouts with conservative value estimation to balance optimism and safety.

Other notable directions include trajectory optimization and decision-based methods, such as Decision Transformer (DT) \cite{chen2021decision} and Implicit Q-Learning (IQL) \cite{kostrikov2021offline}, which cast offline RL as a supervised learning problem over sequences or value distributions. Additionally, imitation-based methods like BAIL \cite{chen2020bail} interpolate between behavior cloning and value-based methods using uncertainty-aware selection of demonstration trajectories. We refer the reader to \cite{levine2020offline, prudencio2023survey} for comprehensive review of offline RL algorithms.

\removed{
Uncertainty modeling plays a central role in offline RL. Epistemic uncertainty, arising from data scarcity or ambiguity in the value function, must be distinguished from aleatoric uncertainty, which stems from inherent environmental randomness. Ensemble-based methods are a practical way to capture epistemic uncertainty. They have been used in both model-based settings (e.g., MOReL \cite{kidambi2020morel}) and model-free methods (e.g., EDAC \cite{an2021uncertainty}) to stabilize learning by regularizing the Bellman backups or penalizing high-variance predictions. However, ensembles can be computationally expensive and coarse. More structured representations of epistemic uncertainty have been proposed using Epistemic Neural Networks (ENNs) (\cite{osband2023epistemic}), which offer a flexible way to encode and sample from belief distributions over value functions. }

While \textbf{uncertainty quantification} is well studied in supervised learning and Bayesian RL \cite{ghavamzadeh2015bayesian}, its structured application in offline reinforcement learning remains underexplored. Traditional methods often conflate epistemic and aleatoric uncertainty or rely on coarse approximations such as ensemble minima, which can misrepresent uncertainty in regions with limited data. Recent work has begun to address these limitations by introducing methods that model epistemic uncertainty more explicitly. For example, \cite{filos2022epistemic} propose Epistemic Value Estimation (EVE), which provides a task-aware mechanism for quantifying value uncertainty in offline settings. Similarly, \cite{shi2022distributionally} explore distributionally robust model-based offline RL using uncertainty sets over dynamics to improve robustness to model misspecification. Other approaches such as \cite{panaganti2022risk} adopt a risk-sensitive view, incorporating epistemic uncertainty directly into policy optimization to avoid unsafe actions. Ensemble-based methods are a practical way to capture epistemic uncertainty. They have been used in both model-based settings (e.g., MOReL \cite{kidambi2020morel}) and model-free methods (e.g., EDAC \cite{an2021uncertainty}) to stabilize learning by regularizing the Bellman backups or penalizing high-variance predictions. Ensemble-based epistemic modeling has also been explored in diffusion-policy frameworks. 
For example, entropy-regularized diffusion policies with Q-ensembles~\cite{zhang2024entropy} leverage ensemble disagreement 
as an uncertainty signal to guide policy sampling toward high-density, reliable regions of the dataset, 
providing a strong empirical demonstration of the benefits of epistemic-aware value estimation in offline RL. However, ensembles can be computationally expensive and coarse. More structured representations of epistemic uncertainty have been proposed using Epistemic Neural Networks (ENNs) \cite{osband2023epistemic}, which offer a flexible way to encode and sample from belief distributions over value functions. Building on these insights, our work introduces a structured, epistemic-robust alternative to ensemble pessimism by defining uncertainty sets over Q-values, allowing richer representations and more targeted conservatism in offline RL.

Additionally, \textbf{benchmarking offline RL} remains challenging due to limited dataset diversity. While D4RL \cite{fu2020d4rl} and RL Unplugged \cite{gulcehre2020rl} have improved standardization, existing benchmarks largely omit risk sensitive evaluation settings. Such behavior policies tend to handle high cost differently depending on whether they are risk averse or risk seeking. This implicit preference skews the data distribution and contributes to epistemic uncertainty, particularly in cases with less data. Despite its significance, there is currently no benchmark that allows systematic control over the risk sensitivity of the behavior policy to study its impact on offline RL performance. Recent work on cross-domain offline RL, such as OTDF~\cite{lyu2025cross}, highlights that even moderate dynamics mismatch 
can significantly degrade offline performance, further motivating controlled data generation and risk-sensitive evaluation protocols. As a first step toward addressing this gap, we introduce a framework that enables controlled variation of behavioral risk preferences using dynamic expectiles. This allows us to generate offline datasets with adjustable risk profiles, facilitating principled evaluation of offline RL algorithms under different uncertainty conditions.  Our proposed framework is aligned with recent efforts like the Minari platform proposed by \cite{minari}, but uniquely focuses on how risk sensitivity shapes epistemic uncertainty in offline datasets.

Building on these insights, this work introduces \textbf{Epistemic Robust Soft Actor-Critic (ERSAC)}, a unified framework for offline RL that models epistemic uncertainty through structured uncertainty sets over Q-values. By replacing ensemble based pessimism with compact and expressive set constructions such as box, convex hull, and ellipsoids, ERSAC enables conservative yet flexible value estimation and policy optimization. We show that SAC-N arises as a special case under box sets, and further extend the framework using \textbf{Epistemic Neural Networks (Epinet)} to construct ellipsoidal uncertainty sets in closed form, reducing runtime without sacrificing performance.

These contributions open several promising directions for future work, including integrating distributional robustness into set construction, incorporating risk-aware objectives, extending epistemic reasoning to multi-agent and hierarchical settings, and establishing theoretical guarantees such as generalization bounds and regret under epistemic uncertainty. Together, our results highlight the potential of structured and efficient epistemic modeling as a foundation for safe, generalizable, and scalable offline reinforcement learning.

\subsection{Machine Replacement example}
\label{appendix:replacement-example}

\begin{table}[h]
    \centering
    \label{tab:expectile_policy}
    \begin{tabular}{lcccccccccc}
        \toprule
        \textbf{\(\tau\)} & \textbf{1} & \textbf{2} & \textbf{3} & \textbf{4} & \textbf{5} & \textbf{6} & \textbf{7} & \textbf{8} & \textbf{9} & \textbf{10} \\
        \midrule
        0.1 & 0 & 0 & 0 & 0 & 0 & 0 & 0 & 0 & 1 & 1 \\
        0.5 & 0 & 0 & 0 & 0 & 0 & 0 & 1 & 1 & 1 & 1 \\
        0.9 & 0 & 0 & 0 & 0 & 0 & 1 & 1 & 1 & 1 & 1 \\
        \bottomrule
    \end{tabular}
    \caption{Optimal actions for each state under different expectile levels \(\tau\). Action 0 corresponds to progressing forward; Action 1 corresponds to jumping to state 1 with -100 reward.}
\end{table}

\begin{figure*}[htbp]
  \centering

  \begin{subfigure}[t]{0.7\textwidth}
    \includegraphics[width=\textwidth]{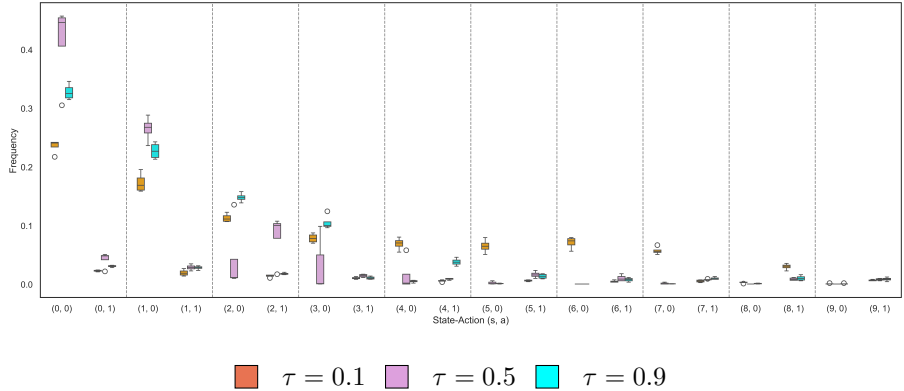}
  \end{subfigure}

  \vspace{1em}

  \begin{tikzpicture}
    \draw[fill=orange] (1,0) rectangle ++(0.3,0.3);
    \node[right] at (1.5,0.15) {\(\tau = 0.1\)};

    \draw[fill=plum] (3,0) rectangle ++(0.3,0.3);
    \node[right] at (3.5,0.15) {\(\tau = 0.5\)};

    \draw[fill=cyan] (5,0) rectangle ++(0.3,0.3);
    \node[right] at (5.5,0.15) {\(\tau = 0.9\)};
  \end{tikzpicture}

  \caption{State visitation frequency distributions under different expectile policies.}
  \label{fig:state_visitation}
\end{figure*}

\subsubsection{Sensitivity of worst-case Q vector to $\pi_\phi$}
\label{appendix:set_comparision}

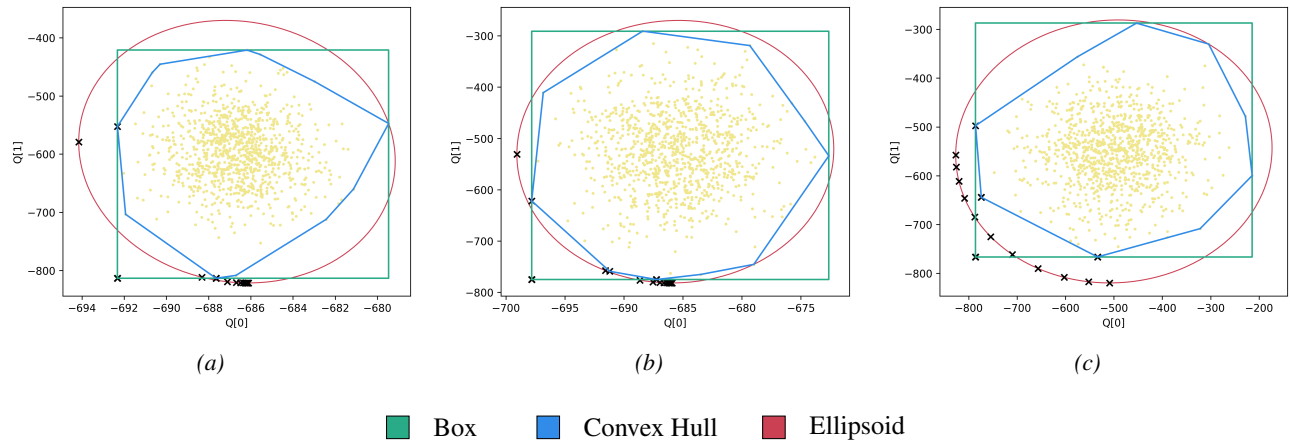
\begin{figure}[htbp]
  \centering

  \begin{subfigure}[t]{0.32\textwidth}
    \includegraphics[width=\textwidth]{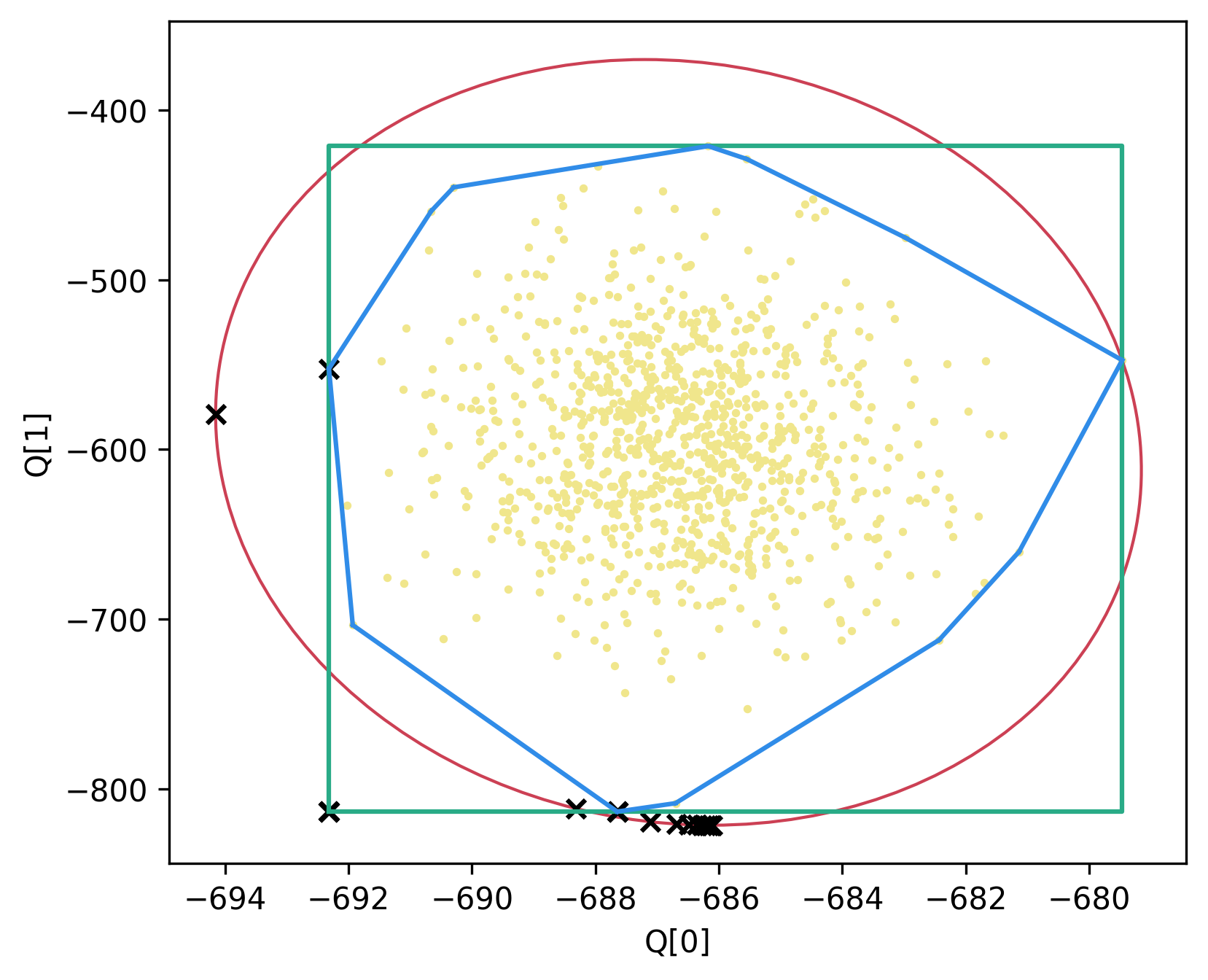}
    \caption{}
  \end{subfigure}
  \hfill
  \begin{subfigure}[t]{0.32\textwidth}
    \includegraphics[width=\textwidth]{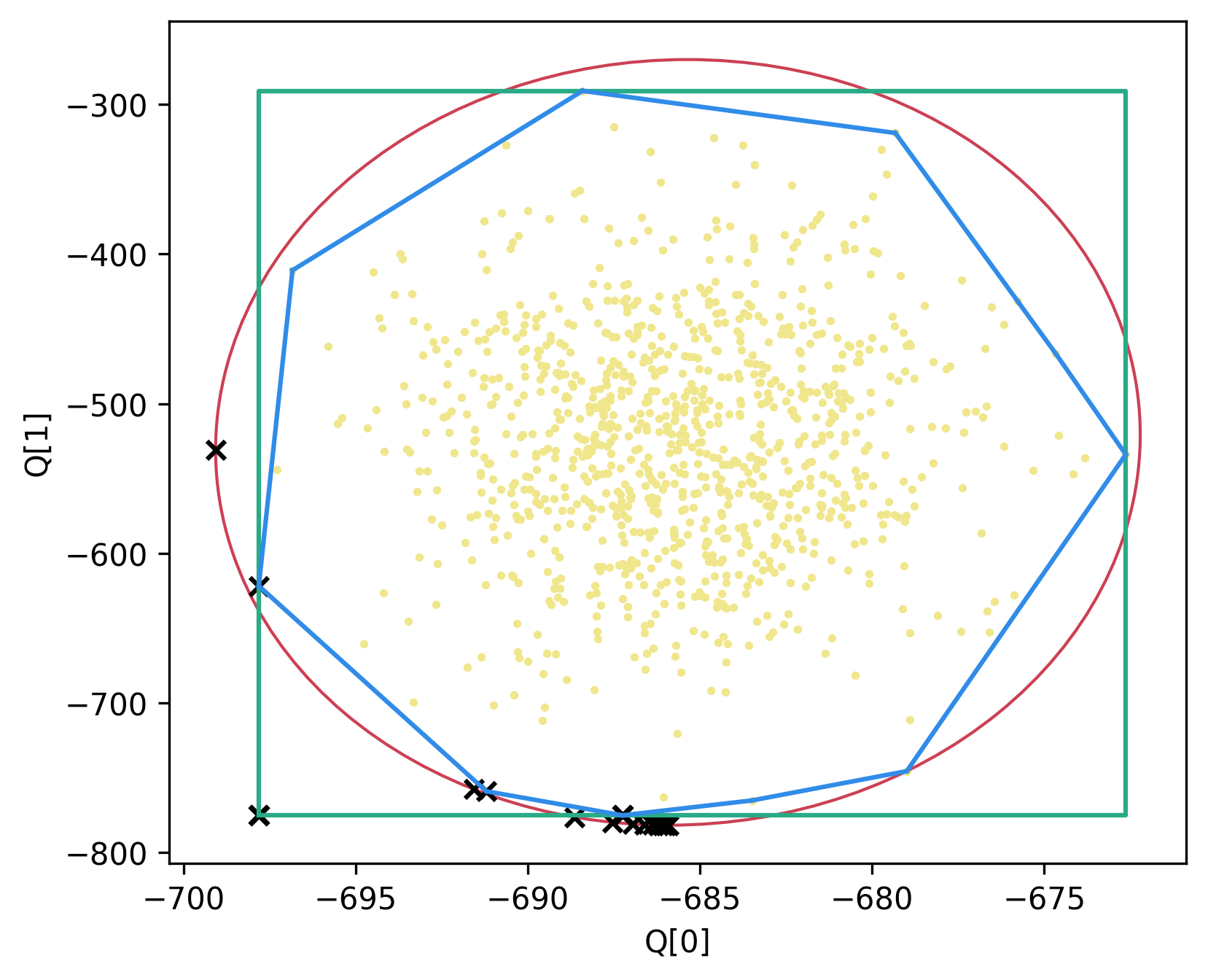}
    \caption{}
  \end{subfigure}
  \hfill
  \begin{subfigure}[t]{0.32\textwidth}
    \includegraphics[width=\textwidth]{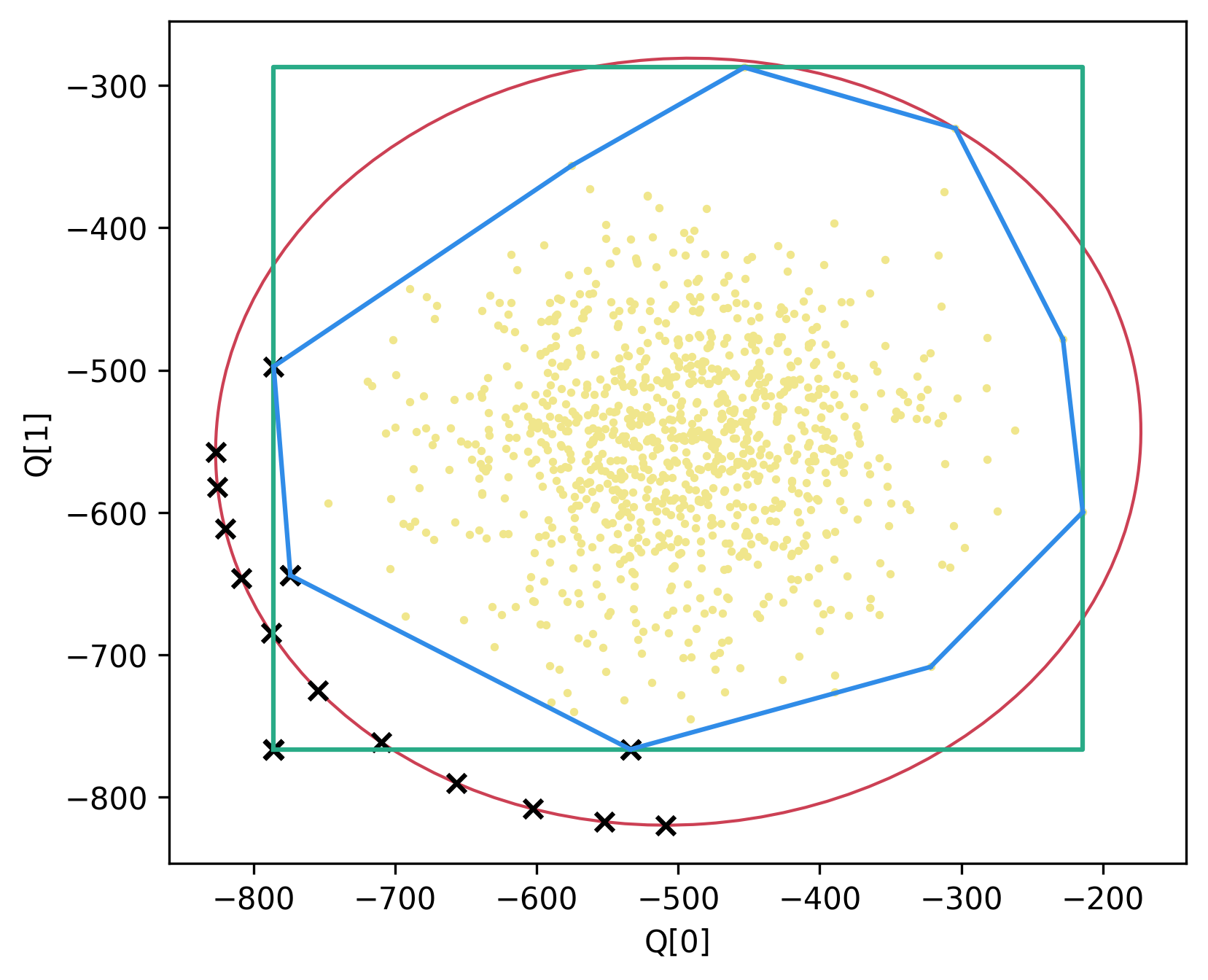}
    \caption{}
  \end{subfigure}

    \vspace{1em}
  \begin{tikzpicture}
    \draw[fill=green] (1,0) rectangle ++(0.3,0.3);
    \node[right] at (1.5,0.15) {Box};

    \draw[fill=blue] (3,0) rectangle ++(0.3,0.3);
    \node[right] at (3.5,0.15) {Convex Hull};

    \draw[fill=red] (6,0) rectangle ++(0.3,0.3);
    \node[right] at (6.5,0.15) {Ellipsoid};
  \end{tikzpicture}

  \caption{(a)--(c): Uncertainty sets and worst-case policy evaluations for states 0, 5, and 10 in the machine replacement example at epoch 1. Each subplot illustrates the distribution of ensemble Q-values along with the corresponding box, convex hull, and ellipsoidal uncertainty sets. Markers X indicate the worst-case Q-value \(q^*\) under different policies \(\pi\).}
  \label{fig:epoch1-selected-states}
\end{figure}

While the box set yields a fixed $q^*(s,\cdot\,;\phi)$ independent of the policy, both the convex hull and ellipsoidal sets adapt their minimizer $q^*(s,\cdot\,; \phi)$ to $\pi_\phi(\cdot \mid s)$. This flexibility introduces a richer learning dynamic, allowing the Bellman backup to respond differently depending on the current policy. This behavior can be viewed from a game-theoretic point of view. At each state $s$, the agent proposes a policy $\pi_\phi(\cdot \mid s)$, and an adversary selects the worst-case Q-vector $q^*(s,\cdot\,; \phi) \in \mathcal{U}_{\theta}(s)$ that minimizes the expected return $\langle \pi_\phi(\cdot \mid s), q \rangle$. When the uncertainty set contains multiple non-dominated extremal points, as is the case for convex hulls and ellipsoids, the Bellman update becomes more responsive capable of adjusting its conservativeness based on the agent's action preferences. To illustrate this, consider the Machine Replacement example discussed above.  Figure \ref{fig:epoch1-selected-states} highlights this adaptivity across selected states by comparing the $q^*$ responses of the three sets $\mathcal{U}_{\text{box}}(s), \mathcal{U}_{\text{hull}}(s)$ and $ \mathcal{U}_{\text{ell}}(s)$  as the policy $\pi$ varies uniformly over the probability simplex. This behavior leads to a more expressive training process that is sensitive to the epistemic structure captured by the generative model.

This adaptivity is particularly important in offline settings, where data coverage is often limited or biased. Structured uncertainty sets enable value estimates that are conservative in underexplored regions while remaining responsive in well-covered ones, leading to improved generalization without excessive pessimism.

The construction of these sets connects with the recent evolving literature in Estimate-then-Optimize Conditional Robust Optimization (CRO). One line of work as proposed in \cite{chenreddy2022data, goerigk2023data, ohmori2021predictive, sun2023predict, blanquero2023contextual} focuses on calibrating uncertainty sets over realizations drawn from a conditional distribution \( F(q \mid s) \). These methods construct high-probability sets \( \mathcal{U}(s) \subset \mathbb{R}^d \) such that for a random realization \( q \sim F(\cdot \mid s) \), it holds that \( \mathbb{P}(q \in \mathcal{U}(s)) \geq 1 - \delta \). Such calibrated sets enable robust decisions of the form $ \max_{\pi \in \Pi} \min_{q \in \mathcal{U}(s)} \pi^\top q,$ that ensure performance against probable realizations of the uncertain quantity \( q \), conditioned on covariates \( s \).

A second line of work, common in distributionally robust optimization and robust RL constructs ambiguity sets over the distribution \( F(\cdot \mid s) \) itself, e.g., using moment constraints, Wasserstein balls, or scenario-based support (\citep{bertsimas2022dynamic, mccord2019phd, wang2020scenario, wang2023learning, nguyen2021robustifying, esteban2022trimmings}). In this setting, one solves:
\[
\max_{\pi \in \Pi} \min_{F \in \mathcal{F}(s)} \mathbb{E}_{q \sim F}[ \pi^\top q ] = \max_{\pi \in \Pi} \min_{\bar{q} \in \mathcal{U}(s)} \pi^\top \bar{q},
\]
where \( \mathcal{F}(s) \) is an ambiguity set over distributions and \( \mathcal{U}(s) := \{ \mathbb{E}_{q \sim F}[q] : F \in \mathcal{F}(s) \} \) is the implied uncertainty set over expected values. 

Our work aligns more closely with the former, wherein we directly parameterize and sample from a learned conditional distribution \( \widehat{F}^q_\theta(s) \), and define a structured uncertainty set \( \mathcal{U}(\widehat{F}^q_\theta(s)) \) over sampled realizations \( q \sim \widehat{F}^q_\theta(s) \). This allows us to reason about epistemic variability in Q-values without requiring a full ambiguity set over \( F^q_\theta(s) \). Bridging these two lines of work could lead to rich formulations for epistemically robust reinforcement learning, which we leave for future work.

\subsection{Proof for Proposition\ref{lem:box_set} }
\label{appendix:sacn-special-case}


\removed{
\begin{lemma}
    When $F_\theta^q(s)$ is defined as \eqref{eq:ensembleF} and $\mathcal{U}(F^q):=\times_{a\in\mathcal{A}} [\essinf_{\tilde{q}\sim F^q}[\tilde{q}(a)],\esssup_{\tilde{q}\sim F^q}[\tilde{q}(a)]]$, then $\mathcal{L}_Q = \sum_i \mathcal{L}_{Q}^i + C$, for some $C\in\mathbb{R}$, and \eqref{eq:policy_loss_robust} reduces to \eqref{eq:ensemble_pi_loss}.
\end{lemma}}
\removed{
   We first show that 
   \[\inf_{q \in \mathcal{U}(F^q_{\theta'}(s'))} \langle \pi_\phi(\cdot \mid s'), q \rangle = \mathbb{E}_{a'\sim\pi_\phi(\cdot \mid s')}[\min_{i \in [N]} Q_{\theta_i}(s',a')]\]}

We begin by analyzing the robust estimator term present in both the conservative target value \eqref{eq:rob_target} and the policy loss \eqref{eq:policy_loss_robust}: $\min_{q \in \mathcal{U}_\theta(s)} \langle \pi_\phi(\cdot \mid s), q \rangle$.
Given that the uncertainty set is defined as a coordinate-wise product box and that $\pi_\phi(\cdot \mid s)\geq 0$, the minimum must be achieved at the coordinate-wise lower bound:
\[
\begin{aligned}
q^*(a) &= \essinf_{\tilde{q} \sim F_\theta^q(s)}[\tilde{q}(a)] \\
&= \essinf_{\tilde{i} \sim U(N)}[Q_{\theta_{\tilde{i}}}(s,a)] \\
&= \min_{i \in [N]} Q_{\theta_i}(s,a), 
\quad \forall a \in \mathcal{A}.
\end{aligned}
\]

The robust evaluation then becomes,
\[
\begin{aligned}
\min_{q \in \mathcal{U}_\theta(s)} \langle \pi_\phi(\cdot \mid s), q \rangle 
&= \sum_{a \in \mathcal{A}} \pi_\phi(a \mid s) 
\min_{i \in [N]} Q_{\theta_i}(s,a) \\
&= \mathbb{E}_{a \sim \pi_\phi(\cdot \mid s)} 
\left[ \min_{i \in [N]} Q_{\theta_i}(s,a) \right]
\end{aligned}
\]

Hence, the conservative target value becomes
\[
\begin{aligned}
y(r, s') &= r + \gamma \, \mathbb{E}_{a' \sim \pi_\phi(\cdot \mid s')} \biggl[
\min_{i \in [N]} Q_{\theta_i}(s', a') 
- \alpha \log \pi_\phi(a' \mid s') \biggr] \\
&= \mathbb{E}_{a' \sim \pi_\phi(\cdot \mid s')} \biggl[
r + \gamma \bigl( 
\min_{i \in [N]} Q_{\theta_i}(s', a') 
- \alpha \log \pi_\phi(a' \mid s') 
\bigr) \biggr] \\
&= \mathbb{E}_{a' \sim \pi_\phi(\cdot \mid s')} 
\bigl[ y(r, s', a') \bigr]
\end{aligned}
\]

We thus have that,

\begin{align*}
    \mathcal{L}_Q^R(\theta)
    &=\mathbb{E}_{\substack{(s,a,r,s') \sim \mathcal{D} \\ \tilde{q} \sim F^q_\theta(s)}}
    \left[ \left( \tilde{q}(a) - y(r,s') \right)^2 \right] \\[0.5em]
    &=\mathbb{E}_{\substack{(s,a,r,s') \sim \mathcal{D} \\ \tilde{q} \sim F^q_\theta(s)}}
    \bigg[
        \tilde{q}(a)^2 
        - 2\tilde{q}(a)\, \mathbb{E}_{a' \sim \pi_\phi(\cdot \mid s')}  \left[y(r,s',a')\right]+\, \mathbb{E}_{a' \sim \pi_\phi(\cdot \mid s')}\left[y(r,s',a')\right]^2
    \bigg] \\[0.5em]
    &=\mathbb{E}_{\substack{(s,a,r,s') \sim \mathcal{D} \\ \tilde{q} \sim F^q_\theta(s)}}
    \bigg[
        \tilde{q}(a)^2 
        - 2\tilde{q}(a)\, \mathbb{E}_{a'}\left[y(r,s',a')\right] +\, \mathbb{E}_{a'}\left[y(r,s',a')^2\right]
    \bigg] \\[0.5em]
    &\qquad +\, \mathbb{E}_{(s,a,r,s') \sim \mathcal{D}}
    \left[  \mathbb{E}_{a'}[y(r,s',a')]^2 - \mathbb{E}_{a'}[y(r,s',a')^2] \right] \\[0.5em]
     \qquad \qquad &=\mathbb{E}_{\substack{(s,a,r,s') \sim \mathcal{D} \\ \tilde{q} \sim F^q_\theta(s) \\ a' \sim \pi_\phi(\cdot \mid s')}}
    \left[ \tilde{q}(a)^2  - 2\tilde{q}(a)\, y(r,s',a') + y(r,s',a')^2 \right] + C \\[0.5em]
    &=\mathbb{E}_{\substack{(s,a,r,s') \sim \mathcal{D} \\ \tilde{q} \sim F^q_\theta(s) \\ a' \sim \pi_\phi(\cdot \mid s')}}
    \left[ \left( \tilde{q}(a) - y(r,s',a') \right)^2 \right] + C \\[0.5em]
    &=\frac{1}{N} \sum_i 
    \mathbb{E}_{\substack{(s,a,r,s') \sim \mathcal{D} \\ a' \sim \pi_\phi(\cdot \mid s')}}
    \left[ \left( Q_{\theta_i}(s,a)  - y(r,s',a') \right)^2 \right] + C \\[0.5em]
    &=\frac{1}{N} \sum_i \mathcal{L}_Q(\theta_i) + C
\end{align*}

where
\begin{align*}
C &:= 
\mathbb{E}_{(s,a,r,s') \sim \mathcal{D}} 
\left[ 
    \left( 
        \mathbb{E}_{a' \sim \pi_\phi(\cdot \mid s')} 
        \left[ y(r,s',a') \right] 
    \right)^2 
\right]  - 
\mathbb{E}_{\substack{(s,a,r,s') \sim \mathcal{D} \\ a' \sim \pi_\phi(\cdot \mid s')}} 
\left[ 
    y(r,s',a')^2 
\right]
\end{align*}

due to $\tilde{q}(a)$ being independent of $y(r,s',a')$ given $(s,a,r,s')$.

On the other hand, we have that:
\begin{align*}
\mathcal{J}_{\pi}^R(\phi)
&= \mathbb{E}_{s \sim \mathcal{D},\ a \sim \pi_\phi(\cdot \mid s)} \bigg[ 
\min_{q \in \mathcal{U}_\theta(s)} 
\langle \pi_\phi(\cdot \mid s), q \rangle 
- \alpha \log \pi_\phi(a \mid s) 
\bigg] \\
&= \mathbb{E}_{s \sim \mathcal{D},\ a \sim \pi_\phi(\cdot \mid s)} \bigg[ 
\mathbb{E}_{a' \sim \pi_\phi(\cdot \mid s)} 
\big[ \min_{i \in [N]} Q_{\theta_i}(s,a') \big] - \alpha \log \pi_\phi(a \mid s) 
\bigg] \\
&= \mathbb{E}_{s \sim \mathcal{D},\ a \sim \pi_\phi(\cdot \mid s)} \bigg[ 
\min_{i \in [N]} Q_{\theta_i}(s,a) - \alpha \log \pi_\phi(a \mid s) \bigg] \\
&= \mathcal{J}_\pi(\phi).
\end{align*}

This completes our proof.

\subsection{Derivations of Worst-Case Q-vector Expressions}
\label{appendix:q_star_derivations}

This section provides derivations supporting the closed-form expressions of the worst-case Q-vector $q^*(s, \cdot; \phi)$ under the convex hull and ellipsoidal uncertainty sets, as referenced in Section~\ref{sec:set_geometry_conservativeness}. These derivations clarify how the worst-case backup depends on the policy $\pi_\phi$.

\subsubsection*{Convex Hull Set}

The worst-case expected Q-value over the convex hull uncertainty set is given by:
\begin{align*}
\min_{q \in \mathcal{U}_{\text{hull}}(\widehat{F}_\theta^q(s))}  
&\mathbb{E}_{a \sim \pi_\phi(\cdot \mid s)} [q(a)] \\
&= \min_{\substack{\lambda \geq 0\\ \sum_{i=1}^N \lambda_i = 1}}  
\mathbb{E}_{a \sim \pi_\phi(\cdot \mid s)} 
\left[ \sum_{i=1}^N \lambda_i\, \mathfrak{q}_\theta(s,a,\tilde{z}_i) \right] \\
&= \min_{\substack{\lambda \geq 0\\ \sum_{i=1}^N \lambda_i = 1}}  
\sum_{i=1}^N \lambda_i\, 
\mathbb{E}_{a \sim \pi_\phi(\cdot \mid s)} 
\left[ \mathfrak{q}_\theta(s,a,\tilde{z}_i) \right] \\
&\geq \min_{i \in [N]} 
\mathbb{E}_{a \sim \pi_\phi(\cdot \mid s)} 
\left[ \mathfrak{q}_\theta(s,a,\tilde{z}_i) \right] \\
&= \mathbb{E}_{a \sim \pi_\phi(\cdot \mid s)} 
\left[ \mathfrak{q}_\theta(s,a,z^*(s,\phi)) \right],
\end{align*}
where $z^*(s,\phi) \in \arg\min_{i} \mathbb{E}_{a \sim \pi_\phi(\cdot \mid s)} \left[ \mathfrak{q}_\theta(s,a,\tilde{z}_i) \right]$.

\subsubsection*{Ellipsoidal Set}

For the ellipsoidal set, we consider the constrained optimization problem:
\begin{align*}
&\min_{q \in \mathcal{U}_{\text{ell}}(\widehat{F}_\theta^q(s))}  
\mathbb{E}_{a \sim \pi_\phi(\cdot \mid s)} [q(a)]\\
&\quad = \min_{\substack{q :\\ (q - \hat{\mu}(s))^\top \widehat{\Sigma}(s)^{-1} (q - \hat{\mu}(s)) \leq \widehat{\Upsilon}(s)^2}}  
\langle \pi_\phi(\cdot \mid s), q \rangle \\
&\quad  = \min_{\substack{\zeta :\\ \|\zeta\| \leq \widehat{\Upsilon}(s)}}  
\langle \pi_\phi(\cdot \mid s), \hat{\mu}(s) + \widehat{\Sigma}^{1/2}(s)\, \zeta \rangle \\
&\quad  \geq \langle \pi_\phi(\cdot \mid s), \hat{\mu}(s) \rangle - \widehat{\Upsilon}(s) \left\| \widehat{\Sigma}^{1/2}(s)\, \pi_\phi(\cdot \mid s) \right\| \\
&\quad  = \left\langle \pi_\phi(\cdot \mid s),\, \hat{\mu}(s) - \widehat{\Upsilon}(s)\cdot \frac{\widehat{\Sigma}(s)\, \pi_\phi(\cdot \mid s)}{\left\| \widehat{\Sigma}^{1/2}(s)\, \pi_\phi(\cdot \mid s) \right\|} \right\rangle,
\end{align*}
where we applied the Cauchy-Schwarz inequality in the third step. This expression matches the closed-form solution for the worst-case Q-vector under the ellipsoidal uncertainty set.

\subsection{Algorithmic Implementation Details}
\label{sec:appendix_algorithms}

In this section, we present the pseudocode for the algorithms discussed in the main paper.

\begin{algorithm}[H]
\caption{Epistemic Robust SAC Training}
\label{alg:robust_sacn}

\begin{algorithmic}[1]
\REQUIRE Initial policy parameters $\phi$, Q parameters $\theta$, target Q parameters $\theta'$, offline replay buffer $\mathcal{D}$, learning rates $\eta_Q, \eta_\pi$, target update rate $\tau$
\ENSURE Updated policy $\phi$ and critic parameters $\theta$

\FOR{each epoch}
    \STATE Sample minibatch $\mathcal{B} := \{(s,a,r,s')\}$ from $\mathcal{D}$

    \STATE \textbf{Compute target:}
    \[
    \begin{aligned}
    y(r, s') \gets\ & r + \gamma \bigg( 
    \min_{q \in \mathcal{U}_{\theta'}(s')} 
    \langle \pi_\phi(\cdot \mid s'),\, q \rangle
    - \alpha\, \mathbb{E}_{a' \sim \pi_\phi}
    [\log \pi_\phi(a' \mid s')] 
    \bigg)
    \end{aligned}
    \]

    \STATE \textbf{Critic update:}
    \[
    \begin{aligned}
    \theta \leftarrow\ & \theta
    - \eta_Q \cdot \tfrac{2}{|\mathcal{B}|} 
    \sum_{(s,a,r,s') \in \mathcal{B}}
    \mathbb{E}_{\tilde{z} \sim F_z} \Big[
    (\mathfrak{q}_\theta(s,a,\tilde{z}) - y(r,s'))
    \cdot \nabla_\theta \mathfrak{q}_\theta(s,a,\tilde{z})
    \Big]
    \end{aligned}
    \]

    \STATE Compute worst-case value vectors:
    \[
    q^*(s,\cdot \,;\phi)
    \leftarrow
    \arg\min_{q \in \mathcal{U}_\theta(s)}
    \langle \pi_\phi(\cdot \mid s),\, q \rangle
    \]

    \STATE \textbf{Actor update:}
    \[
    \begin{aligned}
    \phi \leftarrow\ & \phi
    + \eta_\pi \cdot \tfrac{1}{|\mathcal{B}|}
    \sum_{s \in \mathcal{B}}
    \Big(
    \sum_{a \in \mathcal{A}}
    q^*(s,a\,;\phi)\, \nabla_\phi \pi_\phi(a \mid s)
    - \alpha\, \nabla_\phi
    \mathbb{E}_{a \sim \pi_\phi(\cdot \mid s)}
    [\log \pi_\phi(a \mid s)]
    \Big)
    \end{aligned}
    \]

    \STATE Update target network:
    \[
    \theta' \leftarrow \tau \theta + (1 - \tau)\theta'
    \]
\ENDFOR

\end{algorithmic}
\end{algorithm}

\newpage
\begin{algorithm}[H]
\caption{Sample-based Epistemic Robust SAC with Box (ERSAC-B) and Convex Hull (ERSAC-CH) Sets}
\label{alg:ERSAC-box-hull}

\begin{algorithmic}[1]
\REQUIRE Initial policy parameters $\phi$, Q parameters $\theta$, target Q parameters $\theta'$, offline data buffer $\mathcal{D}$, learning rates $\eta_Q,\eta_\pi$, target update rate $\tau$, sample size $N$
\ENSURE Updated parameters $\theta,\phi$ and target parameters $\theta'$

\FOR{each epoch}
    \STATE Sample minibatch $\mathcal{B} := \{(s,a,r,s')\}$ from $\mathcal{D}$.
    \STATE Sample $N$ i.i.d. latent variables $\{\tilde{z}_i\}_{i=1}^N$ from $F_z$.

    \STATE \textbf{Construct sampled Q-values:}
    \[
      \mathcal{Q}(s) \leftarrow \{ \mathfrak{q}_\theta(s, \cdot, \tilde{z}_i) \}_{i=1}^N,
      \qquad
      \mathcal{Q}(s') \leftarrow \{ \mathfrak{q}_{\theta'}(s', \cdot, \tilde{z}_i) \}_{i=1}^N .
    \]

    \STATE \textbf{Construct robust targets:}

    \STATE \emph{(Box set)}
    \[
    \begin{aligned}
    y_{\text{box}}(r, s') =\ & r + \gamma \bigg(
    \sum_{a \in \mathcal{A}} \pi_\phi(a \mid s')
    \cdot \min_{i \in [N]} \mathfrak{q}_{\theta'}(s', a, \tilde{z}_i)
    - \alpha \sum_{a \in \mathcal{A}} \pi_\phi(a \mid s')
    \log \pi_\phi(a \mid s') \bigg).
    \end{aligned}
    \]

    \STATE \emph{(Convex Hull set)}
    \[
    \begin{aligned}
    y_{\text{hull}}(r, s') =\ & r + \gamma \bigg(
    \min_{i \in [N]} \sum_{a \in \mathcal{A}}
    \pi_\phi(a \mid s') \cdot \mathfrak{q}_{\theta'}(s', a, \tilde{z}_i)
    - \alpha \sum_{a \in \mathcal{A}}
    \pi_\phi(a \mid s') \log \pi_\phi(a \mid s') \bigg).
    \end{aligned}
    \]

    \STATE \textbf{Critic update (common):}
    \[
    \begin{aligned}
    \theta \leftarrow\ & \theta - \eta_Q \cdot \frac{2}{|\mathcal{B}|}
    \sum_{(s,a,r,s') \in \mathcal{B}}
    \mathbb{E}_{\tilde{z} \sim F_z}
    \Big[ \big( \mathfrak{q}_\theta(s,a,\tilde{z}) - y(r,s') \big)
    \cdot \nabla_\theta \mathfrak{q}_\theta(s,a,\tilde{z}) \Big].
    \end{aligned}
    \]

    \STATE \textbf{Actor update:}

    \STATE \emph{(Box set)}
    \[
    \begin{aligned}
    \phi \leftarrow\ & \phi + \eta_\pi \cdot \frac{1}{|\mathcal{B}|}
    \sum_{s \in \mathcal{B}} \Bigg(
    \sum_{a \in \mathcal{A}} \min_{i \in [N]}
    \mathfrak{q}_\theta(s, a, \tilde{z}_i)\,
    \nabla_\phi \pi_\phi(a \mid s)
    - \alpha\, \nabla_\phi
    \mathbb{E}_{a \sim \pi_\phi(\cdot \mid s)}
    \big[ \log \pi_\phi(a \mid s) \big]
    \Bigg).
    \end{aligned}
    \]

    \STATE \emph{(Convex Hull set)}
    \[
      i^* = \arg\min_{i \in [N]}
      \sum_{a \in \mathcal{A}} \pi_\phi(a \mid s)\, \mathfrak{q}_\theta(s, a, \tilde{z}_i).
    \]
    \[
    \begin{aligned}
    \phi \leftarrow\ & \phi + \eta_\pi \cdot \frac{1}{|\mathcal{B}|}
    \sum_{s \in \mathcal{B}} 
    \sum_{a \in \mathcal{A}}
    \mathfrak{q}_\theta(s, a, \tilde{z}_{i^*}) \cdot \nabla_\phi \pi_\phi(a \mid s)
    - \alpha \cdot \nabla_\phi
    \mathbb{E}_{a \sim \pi_\phi(\cdot \mid s)}
    \big[ \log \pi_\phi(a \mid s) \big].
    \end{aligned}
    \]

    \STATE \textbf{Target network update:}
    \[
      \theta' \leftarrow \tau \theta + (1 - \tau)\theta'.
    \]
\ENDFOR

\end{algorithmic}
\end{algorithm}

\newpage
\begin{algorithm}[H]
\caption{Sample-based Epistemic Robust SAC with Ellipsoidal Uncertainty (ERSAC-E)}
\label{alg:ERSAC-E}

\begin{algorithmic}[1]
\REQUIRE Initial policy parameters $\phi$, Q parameters $\theta$, target Q parameters $\theta'$, offline data replay buffer $\mathcal{D}$, learning rates $\eta_Q,\eta_\pi$, target update rate $\tau$, sample size $N$
\ENSURE Updated parameters $\theta,\phi$ and target parameters $\theta'$

\FOR{each epoch}
    \STATE Sample minibatch $\mathcal{B} := \{(s,a,r,s')\}$ from $\mathcal{D}$

    \STATE Sample $N$ i.i.d. realizations $\{\tilde{z}_i\}_{i=1}^N$ from $F_z$

    \STATE Compute:
    \[
      \hat{\mu}(s) \leftarrow \frac{1}{N}
      \sum_{i=1}^N \mathfrak{q}_\theta(s,\cdot,\tilde{z}_i)
    \]
    \[
      \widehat{\Sigma}(s) \leftarrow \frac{1}{N}\!
      \sum_{i=1}^N
      \big(\mathfrak{q}_\theta(s,\cdot,\tilde{z}_i)-\hat{\mu}(s)\big)
      \big(\mathfrak{q}_\theta(s,\cdot,\tilde{z}_i)-\hat{\mu}(s)\big)^\top
    \]
    \[
    \widehat{\Upsilon}(s) \leftarrow
    \inf \Bigg\{ \Upsilon :
    \frac{1}{N} \sum_{i=1}^N
    \mathbf{1}\!\Big[
      (\mathfrak{q}_\theta(s,\cdot,\tilde{z}_i)-\hat{\mu}(s))^\top
      \widehat{\Sigma}(s)^{-1}
      (\mathfrak{q}_\theta(s,\cdot,\tilde{z}_i)-\hat{\mu}(s))
      \le \Upsilon^2
    \Big]
    \ge \upsilon \Bigg\}
    \]
    \[
    \hat{\mu}(s') \leftarrow \frac{1}{N} \sum_{i=1}^N \mathfrak{q}_{\theta'}(s',\cdot,\tilde{z}_i)
    \] 
    \[
    \widehat{\Sigma}(s') \leftarrow \frac{1}{N} \sum_{i=1}^N (\mathfrak{q}_{\theta'}(s',\cdot,\tilde{z}_i)  - \hat{\mu}(s'))(\mathfrak{q}_{\theta'}(s',\cdot,\tilde{z}_i) - \hat{\mu}(s'))^\top
    \]
    \[
    \widehat{\Upsilon}(s')\leftarrow \inf\{\Upsilon| \frac{1}{N} \sum_{i=1}^N \1\{(\mathfrak{q}_{\theta'}(s',\cdot,\tilde{z}_i) - \hat{\mu}(s'))^\top \widehat{\Sigma}(s')^{-1}(\mathfrak{q}_{\theta'}(s',\cdot,\tilde{z}_i) - \hat{\mu}(s'))\leq \Upsilon^2\} \geq \upsilon\}
    \]
    \STATE \textbf{Compute target:}
    \[
    \begin{aligned}
    y(r,s') \gets\ & r + \gamma \Big(
      \langle \pi_\phi(\cdot\mid s'),\,\hat{\mu}(s') \rangle
      - \widehat{\Upsilon}(s')\,
      \big\| \widehat{\Sigma}^{1/2}(s') \pi_\phi(\cdot\mid s') \big\| \\
      &\qquad\qquad
      - \alpha\,\mathbb{E}_{a'\sim\pi_\phi}
      \big[\log\pi_\phi(a'\mid s')\big]
    \Big)
    \end{aligned}
    \]

    \STATE \textbf{Critic update:}
    \[
    \begin{aligned}
    \theta \leftarrow\ & \theta - \eta_Q \cdot \tfrac{2}{|\mathcal{B}|}
    \sum_{(s,a,r,s')\in\mathcal{B}}
    \mathbb{E}_{\tilde{z}\sim F_z}
    \Big[
      (\mathfrak{q}_\theta(s,a,\tilde{z}) - y(r,s'))\,
      \nabla_\theta \mathfrak{q}_\theta(s,a,\tilde{z})
    \Big]
    \end{aligned}
    \]

    \STATE \textbf{Actor update:}
    \[
    \begin{aligned}
    \phi \leftarrow\ & \phi + \eta_\pi \cdot \tfrac{1}{|\mathcal{B}|} \sum_{s \in \mathcal{B}} \sum_{a \in \mathcal{A}} \biggl( 
\hat{\mu}(s, a) 
- 
\widehat{\Upsilon}(s) \cdot
\tfrac{ 
\left[ \widehat{\Sigma}(s) \pi_\phi(\cdot \mid s) \right](a)
}{ 
\left\| \widehat{\Sigma}^{1/2}(s) \pi_\phi(\cdot \mid s) \right\| 
} 
\biggr) \nabla_\phi \pi_\phi(a \mid s)  - \alpha\, \nabla_\phi \mathbb{E}_{a \sim \pi_\phi(\cdot \mid s)} 
\left[ \log \pi_\phi(a \mid s) \right]
    \end{aligned}
    \]

    \STATE Update target network:
    \[
      \theta' \leftarrow \tau \theta + (1-\tau)\theta'
    \]
\ENDFOR

\end{algorithmic}
\end{algorithm}

\begin{algorithm}[H]
\caption{Sample-based ERSAC with Ellipsoidal Uncertainty using Epinet (ERSAC-E-Epi)}
\label{alg:ERSAC-E-Epi}

\begin{algorithmic}[1]
\REQUIRE Initial policy parameters $\phi$; Q-network parameters
$\theta=(\theta_\mu,\theta_\sigma)$; target network parameters
$\theta'=(\theta'_\mu,\theta'_\sigma)$; offline data buffer $\mathcal{D}$;
learning rates $\eta_Q,\eta_\pi$; target update rate $\tau$; noise scale
$\bar{\sigma}$; regularization coefficients $\lambda_\mu,\lambda_\sigma$;
sample size $N$
\ENSURE Updated parameters $\phi,\theta$ and target parameters $\theta'$

\FOR{each epoch}
    \STATE Sample minibatch
    $\bar{\mathcal{B}} := \{(s,a,r,s',c)\}$
    from augmented buffer $\bar{\mathcal{D}}$,
    where $c \sim \mathrm{Unif}(\mathbb{S}^{d_z})$

    \STATE Sample $N$ i.i.d. latent indices
    $\{\tilde{z}_i\}_{i=1}^N \sim \mathcal{N}(0,I)$

    \STATE \textbf{Construct uncertainty set (Epinet ellipsoid):}

    \STATE Compute mean:
    \[
      \hat{\mu}(s') \gets \mu_{\theta'_\mu}(s')
    \]

    \STATE Compute Epinet variance features:
    \[
      \bar{\sigma}_{\theta'}(s',a)
      \gets
      \bar{\sigma}^L_{\theta'_\sigma}\big(\psi_{\theta'_\mu}(s'),a\big)
      + \bar{\sigma}^P\big(\psi_{\theta'_\mu}(s'),a\big)
    \]

    \STATE Construct covariance:
    \[
      \Sigma_{\theta'}(s')_{a,a'}
      \gets
      \big\langle
        \bar{\sigma}_{\theta'}(s',a),
        \bar{\sigma}_{\theta'}(s',a')
      \big\rangle
    \]

    \STATE \textbf{Compute robust target:}
    \[
    \begin{aligned}
    y(r,s') \gets\ & r + \gamma \Big(
      \langle \pi_\phi(\cdot \mid s'), \hat{\mu}(s') \rangle
      - \rho \left\|
        \Sigma_{\theta'}^{1/2}(s') \pi_\phi(\cdot \mid s')
      \right\|_2 
      - \alpha\,
      \mathbb{E}_{a' \sim \pi_\phi}
      [\log \pi_\phi(a' \mid s')]
    \Big)
    \end{aligned}
    \]

    \STATE \textbf{Critic update (mean head):}
    \[
    \begin{aligned}
    \theta_\mu \gets\ &
    \theta_\mu
    - 2\eta_Q \cdot \tfrac{1}{|\bar{\mathcal{B}}|}
    \sum_{(s,a,r,s',c)\in\bar{\mathcal{B}}}
    \mathbb{E}_{\tilde{z}\sim\mathcal{N}(0,I)}
    \Big[
      (\mathfrak{q}_\theta(s,a,\tilde{z})
      - y(r,s')
      - \bar{\sigma}\langle c,\tilde{z}\rangle)
      \nabla_{\theta_\mu}\mu_{\theta_\mu}(s,a)
    \Big]
    + 2\lambda_\mu \theta_\mu
    \end{aligned}
    \]

    \STATE \textbf{Critic update (Epinet head):}
    \[
    \begin{aligned}
    \theta_\sigma \gets\ &
    \theta_\sigma
    - 2\eta_Q \cdot \tfrac{1}{|\bar{\mathcal{B}}|}
    \sum_{(s,a,r,s',c)\in\bar{\mathcal{B}}}
    \mathbb{E}_{\tilde{z}\sim\mathcal{N}(0,I)}
    \Big[
      (\mathfrak{q}_\theta(s,a,\tilde{z})
      - y(r,s')
      - \bar{\sigma}\langle c,\tilde{z}\rangle)
      \nabla_{\theta_\sigma}
      \sigma^L_{\theta_\sigma}
      (\psi_{\theta_\mu}(s),a,\tilde{z})
    \Big]
    + 2\lambda_\sigma \theta_\sigma
    \end{aligned}
    \]

    \STATE \textbf{Actor update:}
    \[
    \begin{aligned}
    \phi \gets\ 
    &\phi + \eta_\pi \cdot \frac{1}{|\bar{\mathcal{B}}|} 
    \sum_{s \in \bar{\mathcal{B}}} \bigg[ 
    \sum_{a\in \mathcal{A}} \bigg( 
    \hat{\mu}(s,a) - \rho \cdot 
    \frac{ \Sigma_\theta(s)\pi_\phi(a \mid s) }
         { \left\| \Sigma_\theta^{1/2}(s)\pi_\phi(\cdot \mid s) \right\| } 
    \bigg) \nabla_\phi \pi_\phi(a \mid s) - \alpha \cdot \nabla_\phi \mathbb{E}_{a \sim \pi_\phi}[ \log \pi_\phi(a \mid s) ] 
    \bigg]
    \end{aligned}
    \]

    \STATE Update target networks:
    \[
      \theta' \gets \tau\,\theta + (1-\tau)\theta'
    \]
\ENDFOR

\end{algorithmic}
\end{algorithm}

\removed{
\subsection{Risk-Sensitive Offline Data Generation}
\begin{algorithm}[H]
\caption{Offline Data Generation via Dynamic Expectile Risk Policies}
\label{alg:expectile_data_gen}
\DontPrintSemicolon
\KwIn{Environment $\mathcal{M}$; risk level $\tau \in (0,1)$; dataset size $|\mathcal{D}|$; learning rates $\eta_Q$, $\eta_\pi$; exploration rate $\epsilon$}
Initialize policy parameters $\phi$ and value function parameters $\theta$ \;\\
\While{not converged}{
    Sample transition $(s, a, r, s')$ by executing current policy $\pi_\phi$ in environment $\mathcal{M}$\;\\
    Compute expectile target:
\EDcomments{I corrected this:
\[     y \gets \arg\min_{z \in \mathbb{R}} \left| \tau - \mathbb{I}(r + \gamma V_\theta(s') < z) \right| \cdot (r + \gamma V_\theta(s') - z)^2     \]
This because in the code I saw:
 \[
    z(r,s') \gets r + \gamma \max_{a'} Q_{\theta'}(s',a')
    \]    
    \[ y : E_{\hat{p}(\cdot|s,a)}[\left| \tau - \mathbb{I}( y < z(r,s')  \right| (y-z(r,s'))]= 0\]
    So
    \[
    y \gets \arg\min_{y \in \mathbb{R}} E_{\hat{p}(\cdot|s,a)}[\left| \tau - \mathbb{I}(y < r + \gamma \max_{a'} Q_{\theta'}(s',a')) \right| \cdot ( y - r - \gamma\max_{a'} Q_{\theta'}(s',a'))^2]
    \]
    We could use:}
    \[
    y \gets \arg\min_{z \in \mathbb{R}} E_{\hat{p}(\cdot|s,a)}[\left| \tau - \mathbb{I}(z < r + \gamma \max_{a'} Q_{\theta}(s',a')) \right| \cdot (z - r - \gamma\max_{a'} Q_{\theta}(s',a'))^2]
    \]
    
    Update value function:
    \[
    \theta \gets \theta - \eta_Q \cdot \nabla_\theta \left(Q_\theta(s,a) - y \right)^2
    \]
    Update policy via policy gradient:
    \EDcomments{Which one?
    \[
    \phi \gets \phi + \eta_\pi \cdot \nabla_\phi \mathbb{E}_{a \sim \pi_\phi(\cdot | s)} [ V_\theta(s) ]
    \]
    \[
    \phi \gets \phi + \eta_\pi \cdot \nabla_\phi \mathbb{E}_{a \sim \pi_\phi(\cdot | s)} [ Q_{\theta}(s,a) ]
    \]
        \[
    \phi \gets \phi + \eta_\pi \cdot  \mathbb{E}_{a \sim \pi_\phi(\cdot | s)} [\nabla_\phi\log(\pi_\phi(a | s))  Q_{\theta}(s,a)  ]
    \]}
}
\textbf{Data Collection with $\epsilon$-Greedy Exploration:} \\
Initialize empty dataset $\mathcal{D} \leftarrow \emptyset$ \;
\While{$|\mathcal{D}| < N$}{
    Observe state $s$ from environment $\mathcal{M}$\;
    With probability $\epsilon$, select action $a \sim \text{Uniform}(\mathcal{A})$\;
    Otherwise, sample action $a \sim \pi_\phi(\cdot | s)$\;
    Execute action $a$ in environment to observe reward $r$ and next state $s'$\;
    Store $(s, a, r, s')$ in buffer $\mathcal{D}$\;
}
\Return Offline dataset $\mathcal{D}$
\end{algorithm}

\EDcomments{
\begin{algorithm}[H]
\caption{Offline Data Generation via Dynamic Expectile Risk Policies ERICK}
\label{alg:expectile_data_gen}
\DontPrintSemicolon
\KwIn{Environment $\mathcal{M}$; risk level $\tau \in (0,1)$; dataset size $|\mathcal{D}|$; learning rates $\eta_Q$, $\eta_\pi$; exploration rate $\epsilon$}
Initialize policy parameters $\phi$ and value function parameters $\theta$ \;
\While{not converged}{
    Sample transition $(s, a, r, s')$ by executing current policy $\pi_\phi$ in environment $\mathcal{M}$\;
    Compute expectile target:
    \[
    y(r,s') \gets r + \gamma \sum_{a'} \pi_\phi (a'|s')Q_{\theta'}(s',a')
    \]
    Update value function:
    \[
    \theta \gets \theta - \eta_Q \cdot \nabla_\theta \left| \tau - \mathbb{I}( Q_\theta(s,a)) < y(r,s')  \right|(Q_\theta(s,a)-y(r,s'))^2
    \]
    \EDcomments{I corrected the use of $\tau$ and am pretty sure this new version is better. Note that the above update reduces to:
    \[
    \theta \gets \theta - 2 \eta_Q \cdot  \left| \tau - \mathbb{I}( Q_\theta(s,a)) < y(r,s')  \right| (Q_\theta(s,a)-y(r,s')) \nabla_\theta Q_\theta(s,a)
    \]
    }
    Update policy via policy gradient:
    \[
    \phi \gets \phi + \eta_\pi \cdot \nabla_\phi \mathbb{E}_{a \sim \pi_\phi(\cdot | s)} [ Q_{\theta'}(s,a) ]
    \]
}
\textbf{Data Collection with $\epsilon$-Greedy Exploration:} \\
Initialize empty dataset $\mathcal{D} \leftarrow \emptyset$ \;
\While{$|\mathcal{D}| < N$}{
    Observe state $s$ from environment $\mathcal{M}$\;
    With probability $\epsilon$, select action $a \sim \text{Uniform}(\mathcal{A})$\;
    Otherwise, sample action $a \sim \pi_\phi(\cdot | s)$\;
    Execute action $a$ in environment to observe reward $r$ and next state $s'$\;
    Store $(s, a, r, s')$ in buffer $\mathcal{D}$\;
}
\Return Offline dataset $\mathcal{D}$
\end{algorithm}
}
}

\subsection{Risk-Sensitive Offline Data Generation}
\label{appendix:data_generation}
\newcommand{\saaN}{{N_{s}}}
\begin{algorithm}[H]
\caption{Offline Data Generation via Dynamic Expectile Risk Policies}
\label{alg:expectile_data_gen}

\begin{algorithmic}[1]
\REQUIRE Environment $\mathcal{M}$; risk level $\tau \in (0,1)$; dataset size $N_\mathcal{D}$; initial policy parameters $\phi$; Q parameters $\theta$; target Q parameters $\theta'$; learning rates $\eta_Q, \eta_\pi$; exploration rate $\epsilon$; number of samples $\saaN$ for $P(\cdot\mid s,a)$ approximation
\ENSURE Offline dataset $\mathcal{D}$

\STATE Initialize policy parameters $\phi$ and value function parameters $\theta$

\FOR{each epoch}
    \STATE Initialize state $s$

    \WHILE{episode not done}
        \STATE Sample transition $(s,a,r,s')$ by executing current policy $\pi_\phi$ in environment $\mathcal{M}$

        \STATE \textbf{Compute expectile target:}
        \[
        \begin{aligned}
        y \gets\ & \sup \bigg\{ z :\
        \mathbb{E}_{s' \sim \hat{p}_{\saaN}(\cdot \mid s,a)} \bigg[
        \left| \tau - \mathbb{I}\left( z < r + \gamma \max_{a'} Q_{\theta'}(s', a') \right) \right|
        \cdot \left( z - r - \gamma \max_{a'} Q_{\theta'}(s', a') \right)
        \bigg] \leq 0
        \bigg\}
        \end{aligned}
        \]
        \STATE where $\hat{p}_{\saaN}(\cdot\mid s,a)$ is the empirical distribution from $\saaN$ resamples of transitions from $(s,a)$

        \STATE \textbf{Update value function:}
        \[
        \theta \gets \theta - \eta_Q \cdot \nabla_\theta \left(Q_\theta(s,a) - y\right)^2
        \]

        \STATE \textbf{Update policy:}
        \[
        \phi \gets \phi + \eta_\pi \cdot
        \mathbb{E}_{a \sim \pi_\phi(\cdot\mid s)}
        \left[ \nabla_\phi \log \pi_\phi(a \mid s)\cdot Q_\theta(s,a) \right]
        \]

        \STATE Move to next state: $s \leftarrow s'$
    \ENDWHILE

    \STATE Update target network:
    \[
      \theta' \leftarrow \tau \theta + (1-\tau)\theta'
    \]
\ENDFOR

\STATE \textbf{Offline Data Collection with $\epsilon$-Greedy Exploration:}

\STATE Initialize empty dataset $\mathcal{D} \leftarrow \emptyset$

\WHILE{$|\mathcal{D}| < N_\mathcal{D}$}
    \STATE Observe state $s$ from environment $\mathcal{M}$

    \IF{$\mathrm{RandomUniform}(0,1) < \epsilon$}
        \STATE Sample action $a \sim \mathrm{Uniform}(\mathcal{A})$
    \ELSE
        \STATE Sample action $a \sim \pi_\phi(\cdot \mid s)$
    \ENDIF

    \STATE Execute action $a$ in environment to observe $r$ and $s'$

    \STATE Store $(s,a,r,s')$ in buffer $\mathcal{D}$
\ENDWHILE

\STATE \textbf{Return} dataset $\mathcal{D}$
\end{algorithmic}
\end{algorithm}

\removed{
\subsection{Experimental Details}
\label{appendix:exp-details}
\subsubsection{Aggregated results by risk level}
\begin{table}[h]
\centering
\begin{tabular}{lccc}
\toprule
\textbf{Env} & \boldsymbol{$\tau = 0.1$} & \boldsymbol{$\tau = 0.5$} & \boldsymbol{$\tau = 0.9$} \\
\midrule
CartPole     & $95 \pm 8$ (1)  & $93 \pm 14$ (2)  & $92 \pm 14$ (3) \\
LunarLander  & $103 \pm 7$ (1) & $99 \pm 5$ (2) & $99 \pm 5$ (2) \\
MR           & $93 \pm 1$ (3)  & $95 \pm 4$ (1)  & $94 \pm 3$ (2) \\
RS           & $87 \pm 15$ (2) & $87 \pm 17$ (1) & $84 \pm 22$ (3) \\
\bottomrule
\end{tabular}
\caption{Aggregated performance of \textbf{Ell\_0.9-N} across environments with mean $\pm$ std and within-environment rank (1 = best).}
\label{tab:ell09n_agg}
\end{table}
}
\subsection{Training algorithm details}
\label{appendix:training_details}
We evaluate all algorithms on a tabular Machine Replacement MDP with $S=10$ states and $A=2$ actions. Transition dynamics are defined probabilistically, with increasing expected costs for continued operation and a reset mechanism triggered by replacement actions. Rewards are state- and transition-dependent, with negative values to simulate maintenance costs and catastrophic penalties for failure. 

To construct behavior policies, we implement risk-sensitive value iteration using the expectile risk measure at levels $\tau \in \{0.1, 0.5, 0.9\}$. Expectile backups are computed by solving a convex root-finding problem for each state-action pair. Policies are derived via one-hot argmax over the resulting Q-values.

We generate offline trajectories using the expectile-optimal policy $\pi_\tau$ for each $\tau$. At each step,  with probability 0.1, a uniformly random action is taken for exploration. We vary the number of transitions $M \in \{100, 1000, 10000\}$ and use ten random seeds per setting. Each trajectory entry records $(s, a, s', r)$.

We evaluate three risk-sensitive SAC-N variants using $N=100$ Q-ensemble members.
Each method includes entropy regularization with coefficient $\alpha = 0.01$ and actor-critic learning rates $\eta_q = \eta_\pi = 0.01$. Target networks are updated using Polyak averaging with $\tau = 0.005$.

We report normalized returns with respect to the optimal and random policies:
\[
\text{Normalized Return} = \frac{V_{\text{eval}} - V_{\text{random}}}{V_{\text{optimal}} - V_{\text{random}}},
\]
averaged over 1000 episodes. Returns are discounted with $\gamma = 0.9$. We repeat all experiments across ten seeds and report the mean and standard deviation.
All code is implemented in Pytorch and NumPy using vectorized operations. Root-finding in expectile computation uses a bisection method with machine epsilon tolerance.

\subsection{Detailed results}
\label{appendix:results}

\begin{figure*}[htbp]
    \centering
    \begin{subfigure}[t]{0.45\textwidth}
        \centering
        \includegraphics[width=\linewidth]{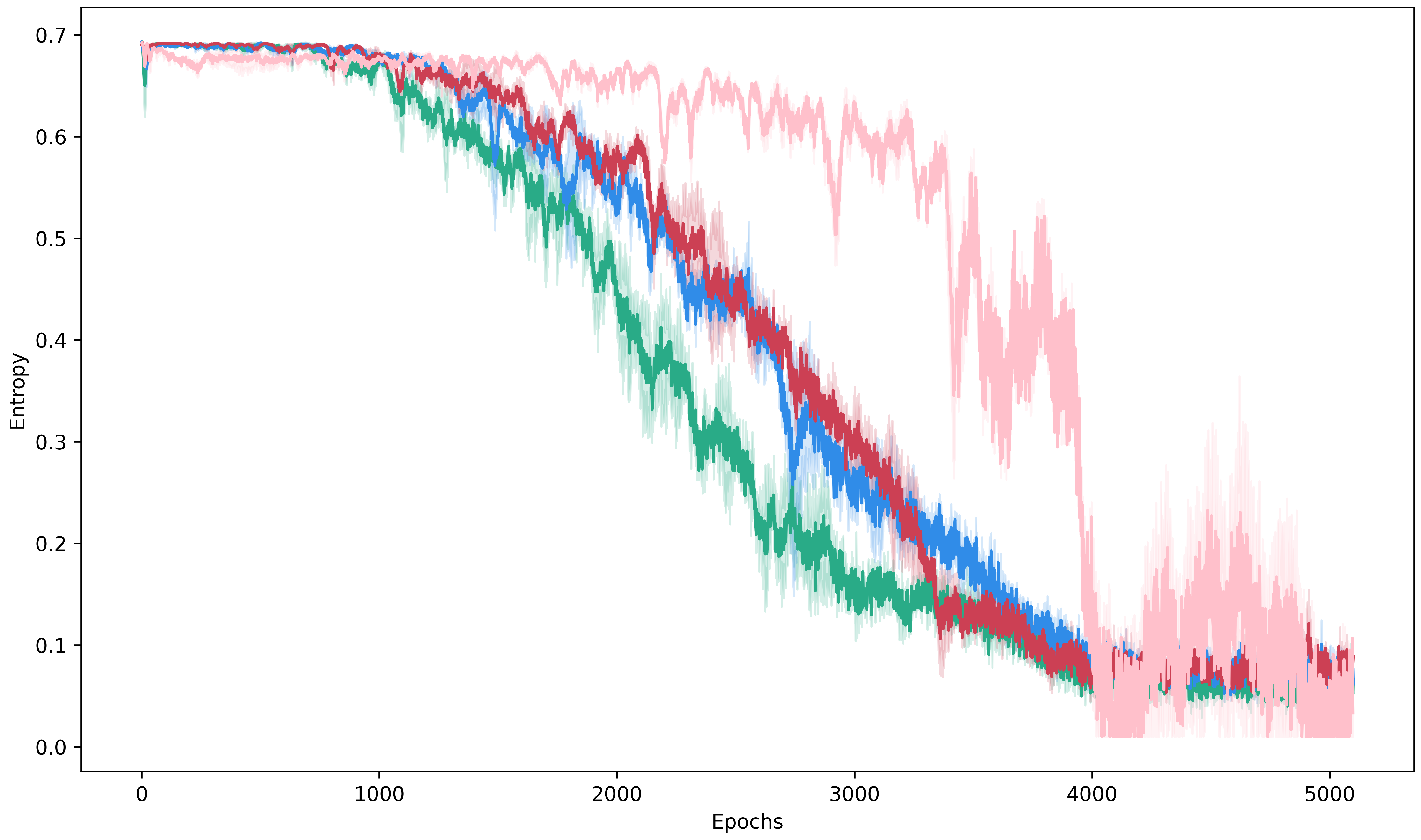}
        \caption{Cartpole}
        \label{fig:cartpole}
    \end{subfigure}
    \hspace{0.01\textwidth}
    \begin{subfigure}[t]{0.45\textwidth}
        \centering
        \includegraphics[width=\linewidth]{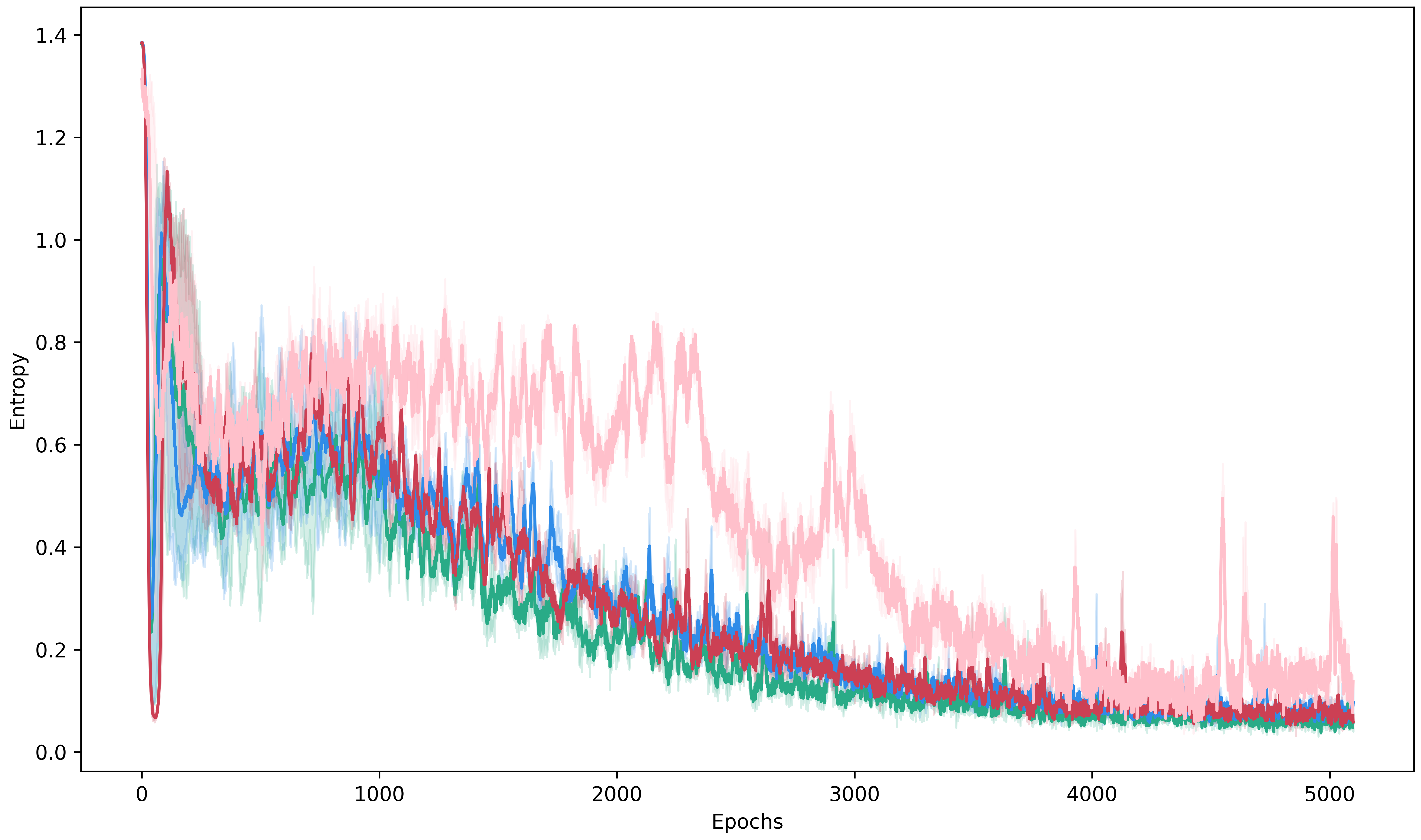}
        \caption{Lunar Lander}
        \label{fig:lunar}
    \end{subfigure}

    \vspace{1em}

    \begin{tikzpicture}
        \matrix[row sep=2pt, column sep=10pt] at (current bounding box.center) {
            \draw[fill=green] (0,0) rectangle ++(0.3,0.3);
            \node[right] at (0.4,0.15) {B\_N}; &

            \draw[fill=blue] (0,0) rectangle ++(0.3,0.3);
            \node[right] at (0.4,0.15) {CH\_N}; &

            \draw[fill=red] (0,0) rectangle ++(0.3,0.3);
            \node[right] at (0.4,0.15) {Ell\_0.9}; &

            \draw[fill=pink] (0,0) rectangle ++(0.3,0.3);
            \node[right] at (0.4,0.15) {Ell\_Epi}; \\
        };
    \end{tikzpicture}

    \caption{Policy entropy during training across B\_N, CH\_N, Ell\_0.9, and Ell\_Epi models in the CartPole and LunarLander environments. Entropy is computed per epoch and averaged over 10 evaluation seeds. Lower entropy indicates more confident, deterministic policies, while higher entropy reflects greater stochasticity in policies.}
    \label{fig:entropy-comparison}
\end{figure*}

This section presents more details about the experiments that are discussed in the main text of the paper. Table \ref{results:tabular_norm_scores} presents additional details on the experiments involving the tabular tasks (i.e., Machine Replacement and RiverSwim). Table \ref{tab:results_gym} presents more detailed statistics about the experiments involving the  CartPole and LunarLander Gym environments. Table \ref{tab:runtime} follows with a report of the runtimes (in s/epoch) of the five offline RL algorithms in the LunarLander Gym. Finally, Figure \ref{fig:entropy-comparison} compares the entropy of the policies obtained from four ER-SAC variants during each epoch of the training. As remarked in the main text, Box-based methods (B-N) maintain consistently lower entropy than  \textbf{CH-N}, \textbf{Ell-N}, and \textbf{Ell-Epi}.

\begin{table*}[t]
  \centering
  \renewcommand{\arraystretch}{1.15}
  \begin{tabular}{lcc|rrrr|r}
    \hline
    \textbf{Env} & \textbf{DS} & \textbf{$\tau$} &
    \textbf{SAC-N} & \textbf{CH-N} & \textbf{Ell-N} & \textbf{Ell\_0.9-N} & \textbf{Beh. Policy} \\
    \hline
    \multirow{9}{*}{\textbf{Machine Replacement}} 
      & 10$\times$   & 0.1 & \underline{$80 \pm 3$} & $85 \pm 2$ & $87 \pm 1$ & $\mathbf{88 \pm 2}$ & $86 \pm 3$ \\
      & 100$\times$  & 0.1 & $97 \pm 1$ & $97 \pm 1$ & \underline{$95 \pm 2$} & $96 \pm 2$ & $86 \pm 3$ \\
      & 1000$\times$ & 0.1 & $98 \pm 2$ & $98 \pm 2$ & $96 \pm 2$ & $96 \pm 1$ & $86 \pm 3$ \\ \cline{2-8}
      & 10$\times$   & 0.5 & \underline{$87 \pm 2$} & $88 \pm 2$ & $90 \pm 2$ & $\mathbf{91 \pm 2}$ & $100 \pm 0$ \\
      & 100$\times$  & 0.5 & $97 \pm 1$ & $\mathbf{98 \pm 1}$ & \underline{$92 \pm 2$} & $94 \pm 2$ & $100 \pm 0$ \\
      & 1000$\times$ & 0.5 & $98 \pm 2$ & $98 \pm 2$ & $98 \pm 2$ & $\mathbf{99 \pm 0}$ & $100 \pm 0$ \\ \cline{2-8}
      & 10$\times$   & 0.9 & \underline{$85 \pm 2$} & $86 \pm 2$ & $90 \pm 2$ & $90 \pm 2$ & $92 \pm 2$ \\
      & 100$\times$  & 0.9 & $96 \pm 2$ & $96 \pm 2$ & \underline{$95 \pm 2$} & $96 \pm 2$ & $92 \pm 2$ \\
      & 1000$\times$ & 0.9 & $96 \pm 2$ & $96 \pm 2$ & $96 \pm 2$ & $96 \pm 1$ & $92 \pm 2$ \\
    \hline
    \multirow{9}{*}{\textbf{RiverSwim}} 
      & 10$\times$   & 0.1 & \underline{$37 \pm 4$} & $64 \pm 2$ & $57 \pm 3$ & $\mathbf{66 \pm 3}$ & $-20 \pm 3$ \\
      & 100$\times$  & 0.1 & \underline{$92 \pm 2$} & $94 \pm 2$ & $94 \pm 3$ & $94 \pm 3$ & $-20 \pm 3$ \\
      & 1000$\times$ & 0.1 & \underline{$99 \pm 1$} & $100 \pm 0$ & $100 \pm 0$ & $100 \pm 0$ & $-20 \pm 3$ \\ \cline{2-8}
      & 10$\times$   & 0.5 & \underline{$56 \pm 2$} & $60 \pm 2$ & $60 \pm 2$ & $\mathbf{62 \pm 1}$ & $100 \pm 0$ \\
      & 100$\times$  & 0.5 & \underline{$97 \pm 2$} & $99 \pm 1$ & $98 \pm 1$ & $99 \pm 1$ & $100 \pm 0$ \\
      & 1000$\times$ & 0.5 & $99 \pm 1$ & $99 \pm 1$ & $100 \pm 0$ & $100 \pm 0$ & $100 \pm 0$ \\ \cline{2-8}
      & 10$\times$   & 0.9 & $49 \pm 2$ & $49 \pm 4$ & \underline{$48 \pm 1$} & $\mathbf{52 \pm 3}$ & $34 \pm 4$ \\
      & 100$\times$  & 0.9 & $99 \pm 1$ & $99 \pm 1$ & $\mathbf{100 \pm 0}$ & $99 \pm 1$ & $34 \pm 4$ \\
      & 1000$\times$ & 0.9 & $99 \pm 1$ & $99 \pm 1$ & $100 \pm 0$ & $100 \pm 0$ & $34 \pm 4$ \\
    \hline
  \end{tabular}
  \caption{Normalized returns with 90\% confidence interval achieved by SAC-N, CH-N, Ell-N, and Ell\_0.9-N across dataset sizes $\{10\times, 100\times, 1000\times\}$ and behavior policy risk levels $\tau \in \{0.1, 0.5, 0.9\}$ in the Machine Replacement and RiverSwim environments. Scores are computed over 10 evaluation seeds and normalized relative to the random and optimal policy baselines. Bold and underline highlight respectively the best and worst performing method when the margin is larger or equal to one. The final column reports the return of the behavior policy used to generate the offline data.}
  \label{results:tabular_norm_scores}
\end{table*}

\begin{table*}[t]
  \centering
  \renewcommand{\arraystretch}{1.15}  
  \begin{tabular}{lcc|rrrrr|r}
    \hline
    \textbf{Env} & \textbf{DS} & \textbf{$\tau$} &
    \textbf{SAC-N} & \textbf{CH-N} & \textbf{Ell\_\!0.9-N} &
    \textbf{Ell-Epi} & \textbf{Ell-Epi$^{*}$} & \textbf{Beh. Policy} \\
    \hline
    \multirow{9}{*}{\textbf{CartPole}}
      & 1k   & 0.1 & $84 \pm 3$ & \underline{$81 \pm 2$} & $\mathbf{86 \pm 1}$
                  & $84 \pm 1$ & $85 \pm 2$ & $86 \pm 2$ \\ 
      & 10k  & 0.1 & \underline{$92 \pm 2$} & $94 \pm 2$ & $100 \pm 0$
                  & $100 \pm 0$ & $100 \pm 0$ & $86 \pm 2$ \\
      & 100k & 0.1 & $100 \pm 0$ & $100 \pm 0$ & $100 \pm 0$
                  & $100 \pm 0$ & $100 \pm 0$ & $86 \pm 2$ \\ \cline{2-9}
      & 1k   & 0.5 & \underline{$70 \pm 2$} & $72 \pm 1$ & $\mathbf{73 \pm 3}$
                  & $72 \pm 2$ & $71 \pm 2$ & $100 \pm 0$ \\
      & 10k  & 0.5 & \underline{$97 \pm 2$} & $99 \pm 1$ & $100 \pm 0$
                  & $100 \pm 0$ & $100 \pm 0$ & $100 \pm 0$ \\
      & 100k & 0.5 & $100 \pm 0$ & $100 \pm 0$ & $100 \pm 0$
                  & $100 \pm 0$ & $100 \pm 0$ & $100 \pm 0$ \\ \cline{2-9}
      & 1k   & 0.9 & $73 \pm 2$ & \underline{$70 \pm 3$} & $78 \pm 2$
                  & $\mathbf{80 \pm 1}$ & $75 \pm 2$ & $83 \pm 2$ \\
      & 10k  & 0.9 & $100 \pm 0$ & $100 \pm 0$ & $100 \pm 0$
                  & $100 \pm 0$ & $100 \pm 0$ & $83 \pm 2$ \\
      & 100k & 0.9 & $100 \pm 0$ & $100 \pm 0$ & $100 \pm 0$
                  & $100 \pm 0$ & $100 \pm 0$ & $83 \pm 2$ \\
    \hline
    \multirow{9}{*}{\textbf{LunarLander}}
      & 1k   & 0.1 & \underline{$72 \pm 1$} & $77 \pm 1$ &
                $98 \pm 2$ & $97 \pm 3$ & $98 \pm 2$ & $94 \pm 3$ \\
      & 10k  & 0.1 & \underline{$94 \pm 2$} & $98 \pm 1$ &
                $102 \pm 1$ & $102 \pm 3$ & $\mathbf{103 \pm 1}$ & $94 \pm 2$ \\
      & 100k & 0.1 & \underline{$99 \pm 1$} & $100 \pm 3$ &
                $106 \pm 1$ & $\mathbf{110 \pm 3}$ & $108 \pm 1$ & $94 \pm 2$ \\ \cline{2-9}
      & 1k   & 0.5 & \underline{$68 \pm 3$} & $73 \pm 3$ &
                $96 \pm 3$ & $95 \pm 1$ & $\mathbf{97 \pm 1}$ & $100 \pm 2$ \\
      & 10k  & 0.5 & \underline{$93 \pm 3$} & $99 \pm 1$ &
                $100 \pm 1$ & $99 \pm 1$ & $\mathbf{102 \pm 1}$ & $100 \pm 2$ \\
      & 100k & 0.5 & \underline{$98 \pm 2$} & $100 \pm 1$ &
                $102 \pm 2$ & $\mathbf{108 \pm 2}$ & $105 \pm 2$ & $100 \pm 2$ \\ \cline{2-9}
      & 1k   & 0.9 & \underline{$67 \pm 2$} & $73 \pm 2$ &
                $97 \pm 2$ & $\mathbf{98 \pm 2}$ & $97 \pm 2$ & $78 \pm 3$ \\
      & 10k  & 0.9 & $92 \pm 2$ & \underline{$92 \pm 3$} &
                $101 \pm 2$ & $100 \pm 4$ & $\mathbf{102 \pm 2}$ & $78 \pm 3$ \\
      & 100k & 0.9 & \underline{$98 \pm 2$} & $101 \pm 2$ &
                $103 \pm 1$ & $104 \pm 2$ & $\mathbf{105 \pm 1}$ & $78 \pm 3$ \\
    \hline
  \end{tabular}
  \caption{Normalized returns with 90\% confidence intervals achieved by the five algorithms across dataset sizes
           $\{1\text{k},10\text{k},100\text{k}\}$ and behavior-policy risk levels
           $\tau\in\{0.1,0.5,0.9\}$ in CartPole and LunarLander.
           Scores are averaged over 10 evaluation seeds and normalized against random and optimal baselines. Bold and underline highlight respectively the best and worst performing method when the margin is larger or equal to one.}
  \label{tab:results_gym}
\end{table*}

\begin{table*}[htbp]
    \centering
    \begin{tabular}{lccccc}
        \toprule
        \textbf{Model} & \textbf{SAC-N} & \textbf{CH-N} & \textbf{Ell\_\!0.9-N} & \textbf{Ell-Epi} & \textbf{Ell-Epi$^{*}$} \\
        \midrule
        Runtime (s/epoch) & 0.35 & 0.42 & 0.56 & 0.60 & 0.10 \\
        \bottomrule
    \end{tabular}
    \caption{Runtime per training epoch for each model in LunarLander with 100{,}000 offline transitions and $\tau = 0.5$, averaged over 10 seeds.}
    \label{tab:runtime}
\end{table*}

\subsection{Additional Experiments on Atari Environments}
\label{appendix:atari}

To evaluate the scalability of ERSAC models to high dimensional observation spaces, we additionally experiment on a subset of Atari 2600 environments from the Arcade Learning Environment (ALE).  
These experiments serve as a test for epistemic robustness in complex domains characterized by pixel based observations, sparse and delayed rewards, and long planning horizons. Unlike the tabular and control settings, we do not introduce risk-sensitive data generation mechanisms in Atari. Instead, we focus on robustness under scale and partial coverage arising from fixed behavior policies.

We evaluate all methods on the following five Atari games,
Breakout, Pong, Q$^{*}$bert, Seaquest, and Hero.
These environments feature high-dimensional pixel observations, sparse or delayed rewards, and long horizons, making them well suited for evaluating robustness under limited coverage. Offline datasets are obtained from the Minari benchmark repository, which provides standardized fixed datasets collected using suboptimal behavior policies.
No additional environment interaction is used during training.
Unlike the earlier tabular and control experiments, we do not vary behavior policy risk sensitivity in the Atari setting.
Instead, these datasets are used to test robustness under scale, partial action coverage, and high dimensional representation learning. Observations follow standard Atari preprocessing: grayscale conversion, frame stacking, and action repeat. All methods use identical convolutional encoders and differ only in the critic and policy objectives.

\textbf{Baselines} We compare the proposed ERSAC variants against several widely used offline reinforcement learning baselines that address extrapolation error and distributional shift through alternative forms of regularization and pessimism. 
Specifically, we evaluate against \textbf{SAC-N}, an ensemble-based Soft Actor-Critic variant that uses pessimistic Bellman backups via the minimum over critics; 
\textbf{Conservative Q-Learning (CQL)}, which enforces conservativeness by regularizing learned action values toward the behavior distribution; 
\textbf{Implicit Q-Learning (IQL)}, which avoids explicit behavior constraints by learning value functions through expectile regression; 
and \textbf{BRAC-BCQ}, a behavior regularized actor critic method that constrains policy updates to remain close to the data distribution. 
These baselines represent state of the art approaches for mitigating overestimation and out of distribution actions in offline reinforcement learning, providing a strong comparison set for evaluating the effectiveness of structured epistemic uncertainty modeling in ERSAC.

For ERSAC, we evaluate ellipsoidal uncertainty sets constructed from ensemble samples (ERSAC-Ell-N) as well as the Epinet-based ellipsoidal variant (ERSAC-Ell-Epi$^{*}$).
For all ellipsoidal methods, the coverage parameter is fixed to $\upsilon = 0.9$, consistent with earlier sections.
Hyperparameters for baseline methods follow published recommendations.

All agents are trained entirely offline for a fixed number of gradient steps per environment. Policies are evaluated deterministically every fixed interval, and final performance is reported as the average episodic return over 100 evaluation episodes. Each reported result is averaged over three random seeds.

\begin{table}[t]
\centering
\renewcommand{\arraystretch}{1.15}
\begin{tabular}{lccccccc}
\toprule
\textbf{Env} & \textbf{SAC-N} & \textbf{CQL} & \textbf{IQL} &
\textbf{BRAC-BCQ} &
\textbf{ERSAC-CH-N} &
\textbf{ERSAC-Ell-N} &
\textbf{ERSAC-Ell-Epi$^{*}$} \\
\midrule
Breakout &
$58 \pm 6$ &
$\mathbf{71 \pm 5}$ &
$68 \pm 4$ &
$35 \pm 7$ &
$62 \pm 5$ &
$64 \pm 5$ &
$66 \pm 4$ \\

Pong &
$78 \pm 8$ &
$\mathbf{86 \pm 6}$ &
$84 \pm 7$ &
$55 \pm 9$ &
$80 \pm 6$ &
$82 \pm 6$ &
$83 \pm 5$ \\

Q*bert &
$54 \pm 7$ &
$63 \pm 6$ &
$\mathbf{66 \pm 6}$ &
$29 \pm 8$ &
$60 \pm 6$ &
$62 \pm 5$ &
$64 \pm 5$ \\

Seaquest &
$33 \pm 6$ &
$47 \pm 5$ &
$45 \pm 5$ &
$18 \pm 6$ &
$42 \pm 6$ &
$46 \pm 5$ &
$\mathbf{49 \pm 5}$ \\

Hero &
$38 \pm 7$ &
$52 \pm 6$ &
$56 \pm 6$ &
$22 \pm 7$ &
$50 \pm 6$ &
$55 \pm 5$ &
$\mathbf{58 \pm 5}$ \\

\bottomrule
\end{tabular}
\caption{
Average episodic returns on Atari~2600 environments using fixed offline datasets from Minari.
Results are reported as mean $\pm$ standard deviation over three random seeds.
ERSAC variants are compared against standard offline RL baselines in high-dimensional visual domains.
}
\label{tab:atari_results}
\end{table}

Across the five Atari environments, ERSAC variants achieve performance comparable to or exceeding ensemble-based SAC-N and remain competitive with specialized offline RL methods such as CQL and IQL.
In environments with sparse rewards and limited effective coverage (e.g., Seaquest and Hero), ellipsoidal ERSAC variants demonstrate more stable learning dynamics than box-based pessimism, suggesting that joint action-level epistemic structure is particularly important in high-dimensional settings.

\end{document}